\PassOptionsToPackage{numbers,square,sort&compress}{natbib}
\PassOptionsToPackage{table,dvipsnames}{xcolor}
\documentclass[10pt]{style/company_light}
\usepackage[utf8]{inputenc}
\makeatletter
\patchcmd{\NAT@sort@cites}
  {\expandafter\immediate\expandafter\write\expandafter\@auxout
      \expandafter{\expandafter\string\expandafter\citation\expandafter{\NAT@cite@list}}}
  {\edef\NAT@tmp@write{\noexpand\write\noexpand\@auxout
      {\noexpand\string\noexpand\citation{\NAT@cite@list}}}\NAT@tmp@write}
  {}{\PackageWarning{main}{Could not patch natbib citation order}}
\makeatother
\usepackage{pifont}
\usepackage{tcolorbox}
\tcbuselibrary{breakable,skins,listings}
\let\origampersand\&
\DeclareRobustCommand{\&}{\textup{\origampersand}}
\newcommand{\workname}{RoboUniverse-10K}
\hypersetup{pdftitle={RoboUniverse-10K: Towards the Scaling Era of Agentic Robotics via Real-to-Sim Environments}}
\title{\workname{}: Toward the Scaling Era of Agentic Robotics via Real-to-Sim Environments}
\author[*,1,2]{Jinkun Liu}
\author[*,1,2]{Yingze Wang}
\author[*,1,2]{Situ Wang}
\author[*,1,2]{Deyao Hong}
\author[*,2,3]{Mingju Gao}
\author[*,2]{Wenyi Li}
\author[*,2]{Yizhe Chi}
\author[1,2]{Youjie Zheng}
\author[1,2]{Kaisen Yang}

\author[1,2]{Xiangyang Wang}
\author[1]{Bingxiang He}
\author[1]{Yuxin Zuo}
\author[1]{Boyuan Chen}
\author[3]{Yue Chen}
\author[4]{Cheng Qian}
\author[5]{Tianxing Chen}
\author[6]{Huichi Zhou}
\author[7]{Huaxiu Yao}
\author[3]{Hao Tang}
\author[1]{Hao Zhou}
\author[1]{Ya-Qin Zhang}

\author[\ddagger,2]{Calvin Xiao}
\author[\ddagger,1,2]{Tiga Altman}
\author[\dagger,\ddagger,1,2]{Huan-ang Gao}

\affiliation[1]{THU}
\affiliation[2]{Einsia.AI}
\affiliation[3]{PKU}
\affiliation[4]{UIUC}
\affiliation[5]{HKU}
\affiliation[6]{UCL}
\affiliation[7]{UNC-Chapel Hill}
\contribution[*]{Equal Contribution}
\contribution[\dagger]{Project Lead}
\contribution[\ddagger]{Corresponding Authors}
\institutionlogo[0.85cm]{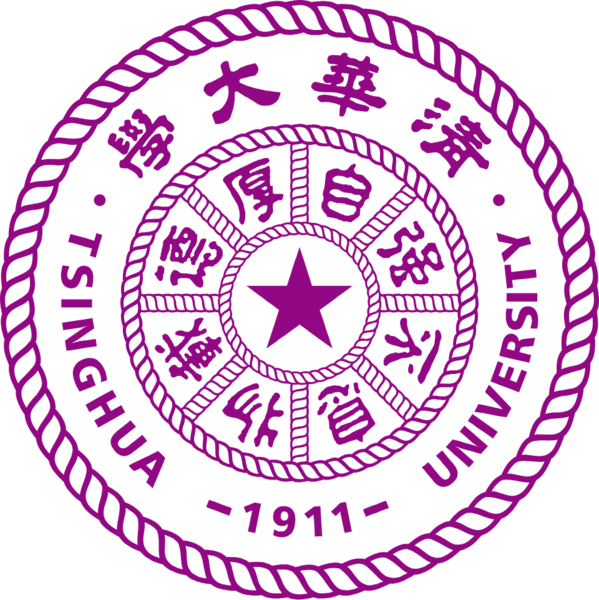}
\institutionlogo[0.85cm]{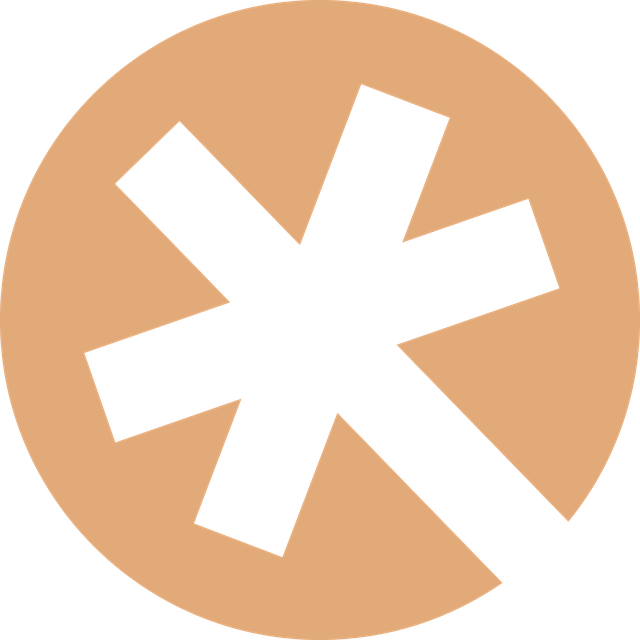}
\institutionlogo[0.85cm]{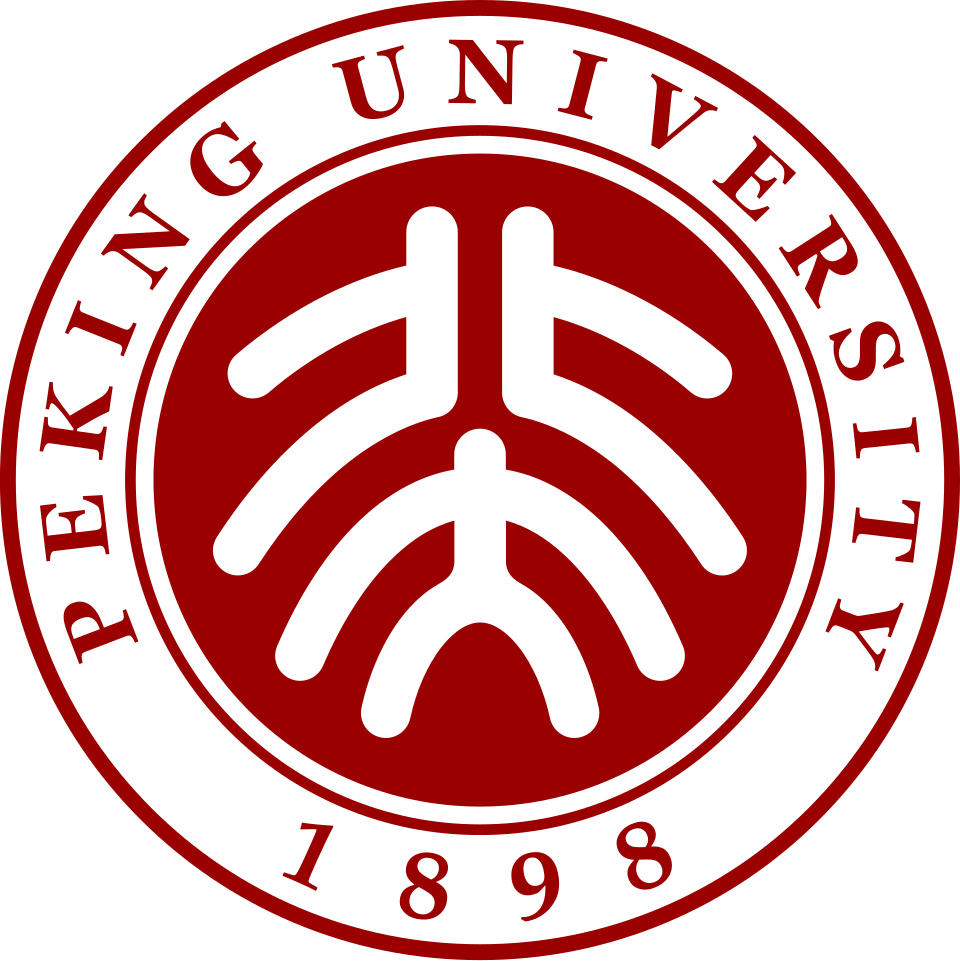}
\institutionlogo[0.85cm]{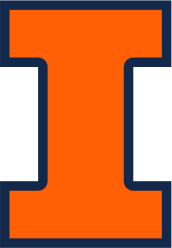}
\institutionlogo[0.85cm]{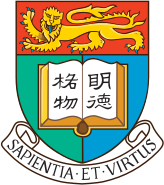}
\institutionlogo[0.85cm]{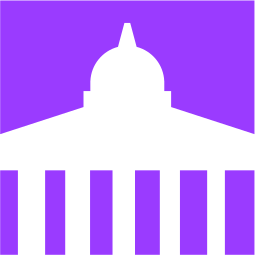}
\institutionlogo[0.85cm]{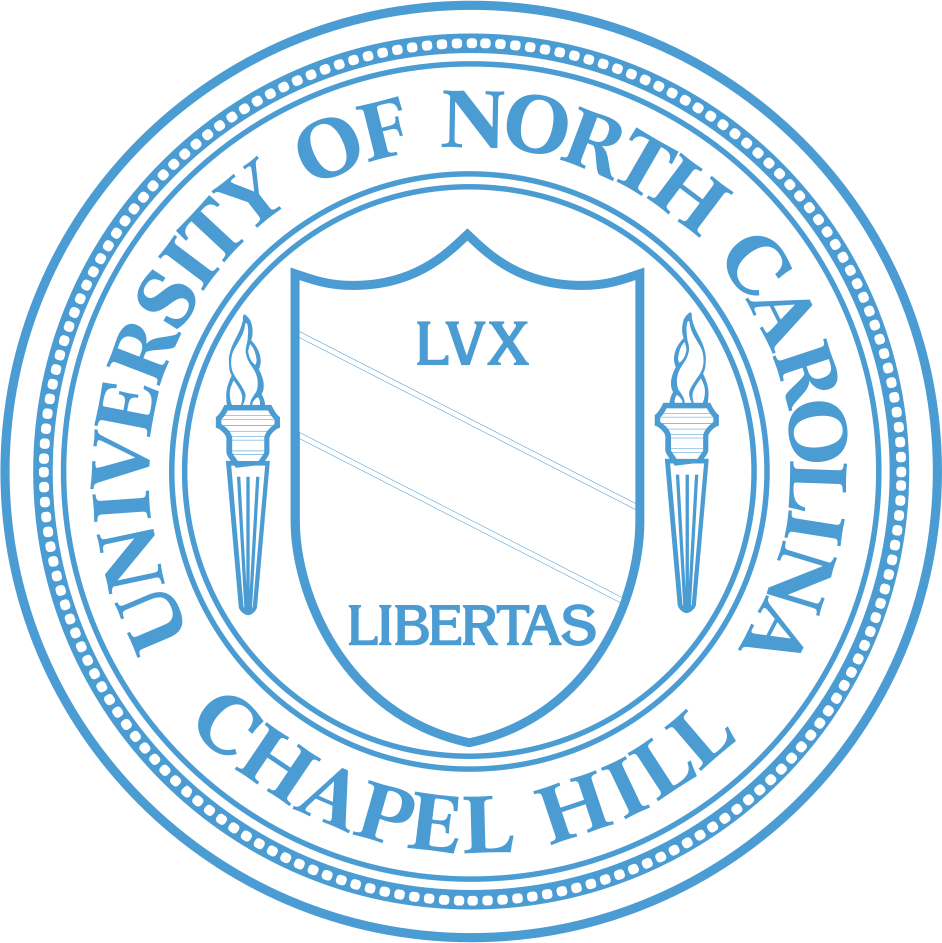}

\metadata[Homepage]{\url{https://lab.einsia.ai/robo-universe}}
\correspondence{\email{gha24@mails.tsinghua.edu.cn}}
\abstract{
Agentic Robotics, in which foundation models act as coding agents to control robots through executable programs, has emerged as a compelling paradigm for embodied intelligence. Yet unlike software engineering, where massive executable environments power post-training, robotics remains constrained by small, manual benchmarks and in-distribution evaluation.
To bridge this gap, we introduce \textbf{RoboUniverse-10K}, 10,807 interactive, resettable, and verifiable physics environments built from real-world videos via an automated Real2Sim and quality assurance pipeline. Each runs in an isolated, Harbor-compatible harness that hides privileged states and grading logic to mirror real-world sensing.
From the same pipeline, we hold out 500 unseen tasks to establish \textbf{RoboUniverse-Verified}, an expert-verified zero-shot benchmark.
Evaluating 11 closed- and open-source models reveals a huge gap: frontier models reach 85.8\% success, while open-source baselines lag at 2.6\% to 10.0\%, bottlenecked by low-level grasping, deformable-object manipulation, and post-contact physical control.
Finally, we collect about 17K multi-turn trajectories totaling 0.79B tokens with GPT-6 Astra and perform full-parameter supervised fine-tuning on Qwen3.6-35B-A3B, yielding \textbf{RoboUniverse-35B-A3B}.
Our model achieves substantial zero-shot gains on RoboUniverse-Verified, lifting the success rate from 0.8\% to 21.2\%, while achieving zero-shot task completions across LIBERO-MAX, RoboCasa, RoboTwin 2.0, and a real Franka arm.
We open-source the RL-ready environments and model to lay the foundation for large-scale reinforcement learning in Agentic Robotics.
}

\date{\today}
\begin{document}
\maketitle
\begin{figure}[h]
    \centering
    \includegraphics[width=\linewidth]{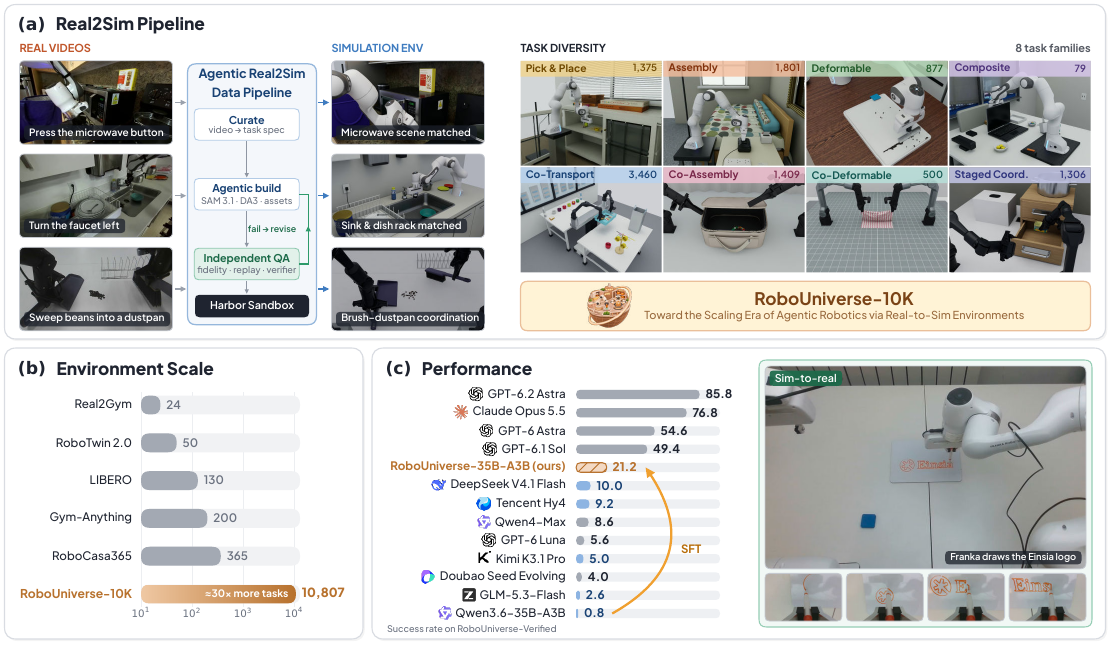}
    \caption{\textbf{Overview of \workname{}.} (a)~An agentic Real2Sim pipeline turns real robot videos into verifiable simulation environments spanning eight task families. (b)~\workname{} is far larger than existing robot simulation benchmarks. (c)~SFT on \workname{} lifts Qwen3.6-35B-A3B from 0.8\% to 21.2\% on RoboUniverse-Verified, and the resulting model transfers zero-shot to a real Franka arm.}
    \label{fig:teaser}
\end{figure}

\section{Introduction}

Recently, \textit{Agentic Robotics}---a paradigm in which frontier foundation models act as \textit{Coding Agents} to control robots---has emerged rapidly. Rather than relying solely on end-to-end visuomotor mappings, these agents write and execute code that invokes perception tools, planning routines, and low-level control interfaces in a closed-loop manner~\citep{liang2022code,huang2023voxposer,fu2026capx,chu2026openrua,yang2026ora,gao2025survey}, already surpassing end-to-end Vision-Language-Action (VLA) policies on several mainstream leaderboards~\citep{robodojoleaderboard,wang2026embodied,openrua2026behind}. Looking to the trajectory of software engineering (SWE), the rapid capability leap of coding agents has been fundamentally driven by \textit{large-scale, executable, and automatically verifiable} post-training environments~\citep{pan2024software,jain2025r2egym,yang2025swesmith,badertdinov2025swerebench,zhao2026immersion,chen2026sweuniverse,chi2026ai4ai}. By mining real-world pull requests to construct hundreds of thousands of interactive repositories, the SWE community unlocked large-scale trajectory distillation and reinforcement learning with verifiable rewards (RLVR)~\citep{wei2025swerl,luo2025deepswe,hong2026swe}, yielding substantial gains for open-source models on benchmarks such as SWE-bench~\citep{jimenez2023swebench}. We posit that the cornerstone of general-purpose Embodied Intelligence is generalization, and the root of generalization lies in the scale and diversity of training environments.

Yet Agentic Robotics has not truly entered its scaling era, primarily due to two fundamental bottlenecks. \textit{First, the bottleneck of environment scale and construction cost.} Existing robotic simulation benchmarks---such as LIBERO, RoboCasa, RoboTwin, and RoboDojo~\citep{liu2023libero,nasiriany2024robocasa,nasiriany2026robocasa365,chen2025robotwin,chen2026robodojo}---rely heavily on manual scene authoring and handcrafted success criteria, keeping their task counts limited to tens or a few hundred. Even recent efforts that leverage LLMs or real-world videos to synthesize simulation tasks (e.g., Real2Gym and Gym-Anything for Robotics)~\citep{wang2023gensim,wang2023robogen,ye2025video2policy,tang2026video2world,ren2026real2gym,aggarwal2026gymanythingrobotics,shen2026embodiedswe,zhang2026pyruadatafarm} remain modest in scale. Unlike software engineering tasks that can be mined directly from existing code repositories, each robotic task requires constructing a physically plausible interactive scene and authoring rigorous verification logic, incurring a far higher construction cost. \textit{Second, the lack of generalization caused by ``train on test.''} Constrained by limited environment scale, prior works~\citep{fan2026captain,liu2026agenticroborl,tang2026reasoning,huang2026learning} typically train and evaluate agents within the same task settings and environment distributions. Training and testing in homogeneous environments not only risks overfitting, but also fails to faithfully demonstrate an agent's true generalization to unseen physical scenes and novel manipulation tasks.

To address these challenges, we introduce \textbf{RoboUniverse-10K}, a suite of 10,807 interactive, resettable, and automatically verifiable Isaac Sim~\citep{nvidia2026isaacsim} environments across 4,132 single-arm Franka and 6,675 dual-arm AgileX tasks for Agentic Robotics post-training and evaluation. Inspired by mining SWE tasks from real-world pull requests, we convert diverse \textit{real-world robot manipulation videos} into distinct concrete tasks via hierarchical deduplication and normalization. Specifically, an \textit{end-to-end Real2Sim construction and quality assurance (QA) pipeline} equips a coding agent with SAM 3.1~\citep{carion2025sam3}, DA3~\citep{lin2025depthanything3}, a curated asset library, and physics-modeling skills to reconstruct scenes, author reference solutions, and write verifiers, which an independent QA agent audits for physical fidelity, reset consistency, and verifier robustness against reward hacking. Meanwhile, a \textit{Harbor-compatible isolated harness} separates the code-execution sandbox from the simulator and verifier, restricting interaction to public observations and control APIs while hiding ground-truth states and grading logic to mirror real-world constraints and support parallel rollouts.

From the same pipeline, we hold out 500 unseen, expert-verified tasks (200 single-arm and 300 dual-arm) across all eight task families to establish the zero-shot benchmark \textbf{RoboUniverse-Verified}, on which we systematically evaluate 11 leading closed- and open-source models.
Our evaluation reveals two core findings. First, a sharp capability gap persists: frontier closed-source models such as GPT-6.2 Astra and Claude Opus 5.5 reach 85.8\% and 76.8\% overall success, whereas open-source baselines lag at 2.6\% to 10.0\% and struggle sharply on dual-arm coordination. Second, failures stem predominantly from physical execution rather than task comprehension. Weaker models are bottlenecked by low-level grasping, dropping to 0\% to 4\% success on grasp-required tasks. Even frontier models struggle with deformable-object manipulation, where cloth tasks barely benefit from capability gains, and roughly 80\% of their failures occur after contact, pointing to fine-grained post-contact physical control as the primary remaining bottleneck.

To validate the utility of RoboUniverse-10K as a scalable post-training infrastructure, we conduct supervised post-training and generalization experiments. Using GPT-6 Astra~\citep{openai2026gpt6astra}, we collect about 17K multi-turn interaction trajectories across the training environments, totaling 0.79B tokens, and perform full-parameter supervised fine-tuning (SFT) on the open-source Qwen3.6-35B-A3B model, yielding \textbf{RoboUniverse-35B-A3B}.
SFT yields substantial zero-shot gains on RoboUniverse-Verified, lifting the overall success rate from 0.8\% to 21.2\% (single-arm from 2.0\% to 40.5\% and dual-arm from 0.0\% to 8.3\%). 
Furthermore, zero-shot probes on LIBERO-MAX, RoboCasa, and RoboTwin 2.0 in the Embodied First Exam~\citep{wang2026embodied} format, alongside a sim-to-real test on a real Franka arm, achieve zero-shot task completions beyond our environments.

In summary, our main contributions are threefold:
\begin{itemize}
    \item \textbf{RoboUniverse-10K:} We present an automated Real2Sim construction and agent-based quality assurance pipeline, and open-source 10,807 Harbor-compatible, isolated, RL-ready simulation environments, overcoming the scale bottleneck limitation of prior benchmarks.
    \item \textbf{RoboUniverse-Verified Benchmark:} We curate an expert-verified zero-shot test suite of 500 tasks, benchmark 11 closed- and open-source models, and pinpoint low-level grasping, deformable-object manipulation, and post-contact physical control as the primary bottlenecks of current embodied coding agents.
    \item \textbf{RoboUniverse-35B-A3B Model:} We train and open-source an SFT model on \workname{}, lifting zero-shot success on RoboUniverse-Verified from 0.8\% to 21.2\% while achieving zero-shot task completions across external benchmarks and a real Franka arm.
\end{itemize}

~\section{The \workname{} Collection}
\label{sec:collection_overview}

\subsection{Environment Design and Execution Harness}
\label{sec:rl_env}

\workname{} provides interactive, resettable, and automatically verifiable physics environments for post-training and evaluating embodied coding agents. Each environment comprises four components: a natural-language task instruction, an Isaac Sim~\citep{nvidia2026isaacsim} physical scene, a robot observation and control interface, and a task-specific verifier. Together, they form a closed loop supporting fast resets and parallel rollouts for both large-scale trajectory collection (SFT) and online reinforcement learning (RL).

\begin{figure*}[t]
    \centering
    \includegraphics[width=\linewidth]{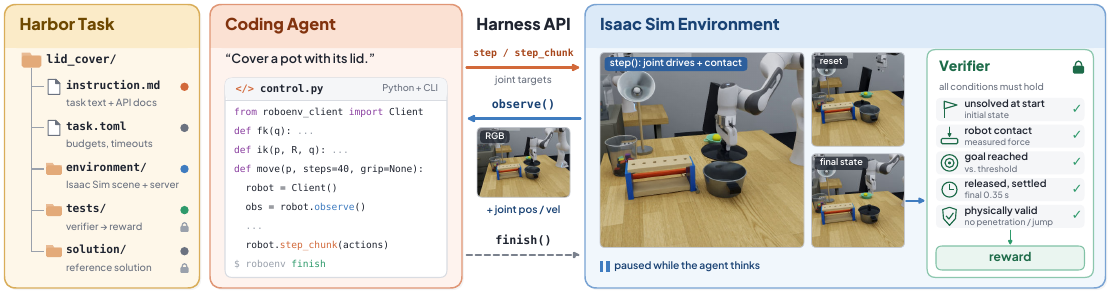}
    \caption{\textbf{A \workname{} environment.} Each Harbor task packages the instruction, configuration, simulation environment, verifier, and reference solution. The coding agent acts only through the harness API, and a hidden verifier checks the final simulator state to return a binary reward.}
    \label{fig:rl_env}
\end{figure*}

\paragraph{Agent Harness and Sandbox Tools.}
We package each task in a format compatible with Harbor~\citep{harbor2026harbor}, the containerized evaluation framework used in Terminal-Bench~\citep{merrill2026terminalbench}, connecting the agent's code-execution sandbox to the simulation backend via a unified harness (\Cref{fig:rl_env}). Upon initialization, the runtime loads the scene and provides the agent with the task instruction and API documentation. Inside the sandbox terminal, the agent writes and runs Python programs that interact with the simulator exclusively through five exposed tools (\Cref{tab:sandbox_tools}). The simulation backend executes joint actuation and physical contact at the specified control frequency and returns updated observations. Simulation time advances only during \texttt{step} and \texttt{step\_chunk} calls, remaining paused during observation reads, code authoring, and model inference.

\paragraph{Execution--Verification Isolation.}
To mirror real-world robotic sensing constraints, the agent sandbox is strictly isolated from the simulation backend and grading logic. Ground-truth object states, reference solutions, and verifiers remain hidden on the environment side. Each episode permits a single attempt bounded by action-step and wall-clock budgets; the agent cannot reset the scene or query intermediate scores. When the agent calls \texttt{finish()} or exhausts its budget, the independent verifier inspects the execution logs and final simulator state to assign a terminal binary reward. By withholding success flags, progress signals, and privileged states, the harness requires the agent to judge physical outcomes and termination timing solely from sensor observations.

\begin{table}[h]
\centering
\footnotesize
\caption{\textbf{Tools exposed by the sandbox to the coding agent.}}
\label{tab:sandbox_tools}
\begingroup
\setlength{\tabcolsep}{2pt}
\begin{tabular*}{\linewidth}{@{\extracolsep{\fill}}l l l c@{}}
\toprule
\textbf{Tool} & \textbf{Input} & \textbf{Output \& Function} & \textbf{\begin{tabular}[b]{@{}c@{}}Advances\\Sim\end{tabular}} \\
\midrule
\texttt{describe()} & --- & Returns action space, observation specs, control period, and budgets. & No \\
\texttt{observe()} & --- & Returns camera images, joint positions/velocities, sim time, and step count. & No \\
\texttt{step(action)} & Single action & Executes one control step and returns updated observations. & Yes \\
\texttt{step\_chunk(actions)} & Action sequence & Executes a sequence of steps and returns the final observation. & Yes \\
\texttt{finish()} & --- & Ends the episode and submits logs to the independent verifier. & No \\
\bottomrule
\end{tabular*}
\endgroup
\end{table}

\subsection{Dataset Composition and Task Taxonomy}

\begin{figure}[t]
    \centering
    \includegraphics[width=\linewidth]{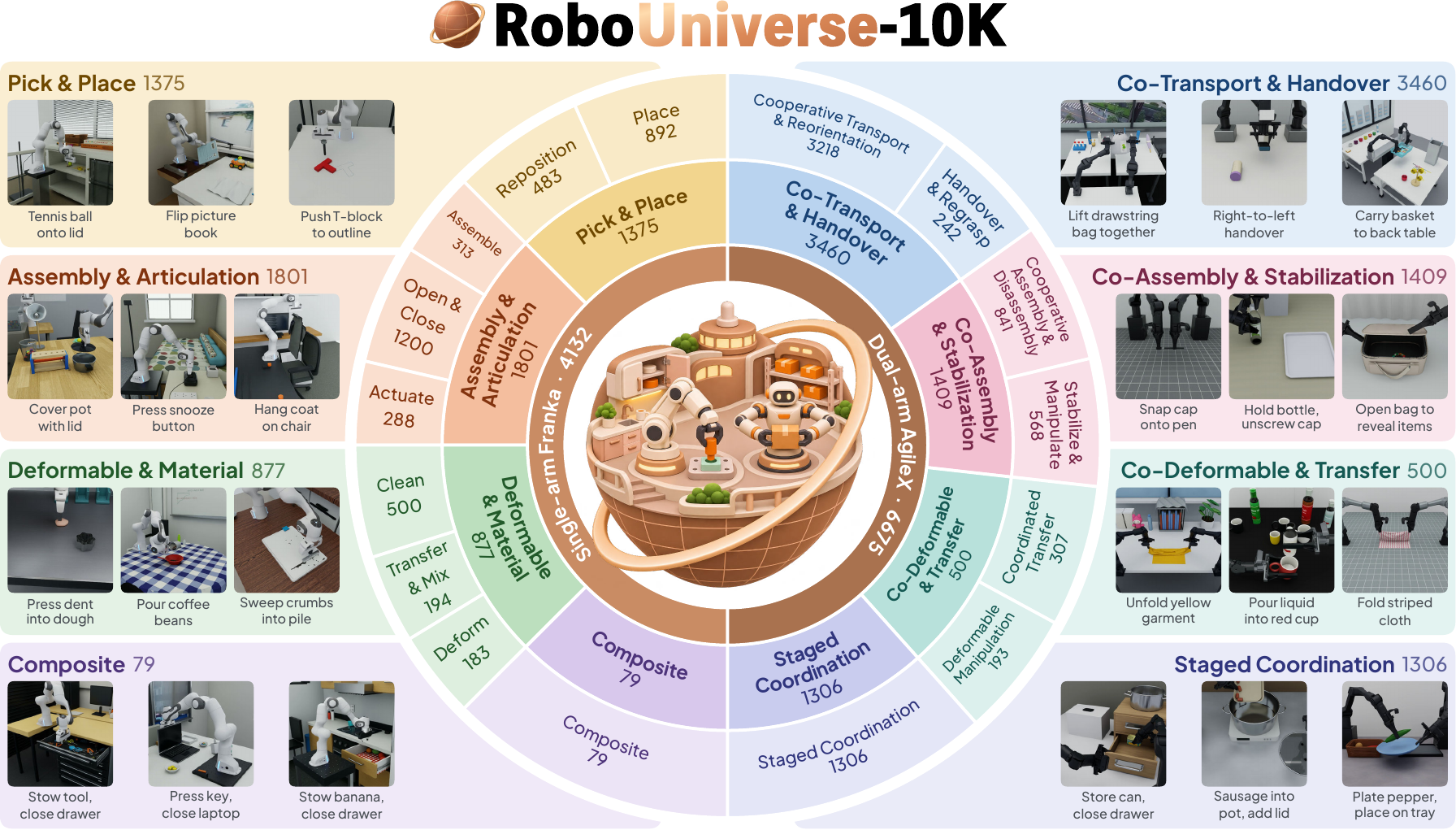}
    \caption{\textbf{Task taxonomy of \workname{}.} The 4,132 single-arm Franka tasks and 6,675 dual-arm AgileX tasks span eight task families. Numbers are task counts; images show example tasks.}
    \label{fig:collection_taxonomy}
\end{figure}

\paragraph{Data Sources and Scale.}
\workname{} comprises 10,807 distinct simulation environments, split into 10,307 training tasks and 500 held-out zero-shot evaluation tasks (RoboUniverse-Verified). DROID~\citep{khazatsky2024droid} is the primary source of demonstrations for the single-arm Franka tasks. We also reconstruct tasks from demonstrations in BridgeData V2~\citep{walke2023bridgedata}, BC-Z~\citep{jang2021bcz}, RT-1~\citep{brohan2023rt1}, Berkeley Fanuc~\citep{zhu2023fanuc}, Berkeley Cable Routing~\citep{luo2024cablerouting}, RoboCook~\citep{shi2023robocook}, Berkeley Autolab UR5~\citep{chen2023berkeleyur5}, Columbia PushT~\citep{chi2023diffusionpolicy}, RoboTurk~\citep{mandlekar2019roboturk}, and CMU Stretch~\citep{bahl2023affordances}. The dual-arm AgileX tasks draw on bimanual teleoperation datasets including RoboDesk~\citep{meituanroboticsteam2026mr0}, RoboCOIN~\citep{wu2025robocoin}, GM100~\citep{wang2026gm100}, RoboDojo~\citep{chen2026robodojo}, ViFailback~\citep{zeng2025vifailback}, MM-30~\citep{liang2026mmabc}, and RoboMIND2~\citep{hou2025robomind2}. Detailed task counts across embodiments and categories are shown in \Cref{fig:collection_taxonomy}.

\paragraph{Hierarchical Deduplication and Normalization.}
Multi-source robotics datasets often exhibit heavy redundancy and inconsistent task granularity. To reduce overlap while preserving task diversity, we organize demonstrations into a five-level hierarchy: \textit{Embodiment} $\rightarrow$ \textit{Taxonomy} $\rightarrow$ \textit{Concrete Task} $\rightarrow$ \textit{Layout} $\rightarrow$ \textit{Trajectory}. We normalize task granularity at the \textit{Concrete Task} level, treating a demonstration as distinct only when it introduces a different interaction type, a substantially different scene context, or a meaningful change in the manipulated objects. Demonstrations differing solely in camera viewpoint, surface appearance, minor spatial arrangements, initial object poses, or execution trajectories are treated as duplicates, with a single representative retained for each concrete task.

\paragraph{Task Taxonomy.}
Based on end-effector contact mechanics and bimanual coordination patterns, we organize the collection into two embodiment tracks covering eight task families, as shown in \Cref{fig:collection_taxonomy}. Single-arm Franka tasks comprise four families. \emph{Pick \& Place} involves rigid-body relocation and 3D reorientation, while \emph{Assembly \& Articulation} covers constrained assembly and sustained-contact interaction with mechanisms such as drawers, doors, and buttons. \emph{Deformable \& Material} addresses shaping non-rigid objects like cloth and dough alongside wiping, pouring, and sweeping, and \emph{Composite} chains multiple primitives within a single episode. Dual-arm AgileX tasks likewise span four families. \emph{Co-Transport \& Handover} covers joint lifting, relocation, and inter-arm passing or regrasping, whereas \emph{Co-Assembly \& Stabilization} focuses on bimanual assembly and asymmetric stabilization and manipulation. \emph{Co-Deformable \& Transfer} involves coordinated folding, flattening, and container-stabilized pouring, and \emph{Staged Coordination} captures asynchronous, causally ordered multi-stage workflows.

\begin{figure}[t]
    \centering
    \includegraphics[width=\linewidth]{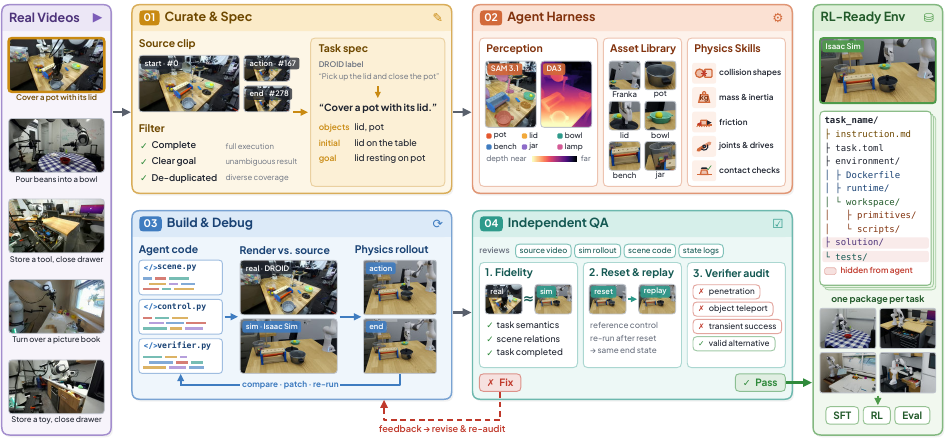}
    \caption{\textbf{Real-to-Sim Data and Quality Assurance Pipeline.} From each real video, a coding agent writes a task specification and builds the scene, reference controller, and verifier with perception tools, an asset library, and physics skills. An independent QA agent checks fidelity, reset and replay, and the verifier, and sends failing environments back for revision. Accepted environments are packaged as Harbor tasks.}
    \label{fig:pipeline}
\end{figure}

\subsection{Real-to-Sim Data and Quality Assurance Pipeline}
\label{sec:pipeline}

\workname{} transforms real-world robot manipulation videos into executable, verifiable RL environments through an automated Real2Sim construction and quality-assurance (QA) pipeline.

\paragraph{Data Curation and Task Specification.}
We first filter real-world manipulation videos to retain episodes with complete executions and clear goals, removing duplicate or near-duplicate scenes while maintaining broad coverage across task families. A coding agent then inspects each video to identify the manipulated objects, initial state, and desired outcome, formulating a natural-language task instruction. Each specification preserves the core objective and necessary constraints of the real demonstration while permitting any valid strategy to complete the task.

\paragraph{Harness-Assisted Environment Construction.}
We equip the coding agent with a unified Real2Sim harness integrating SAM 3.1~\citep{carion2025sam3} segmentation, DA3~\citep{lin2025depthanything3} depth estimation, a pre-built asset library, and Isaac Sim. Using segmentation masks and depth maps to recover object contours and spatial layouts, the agent selects robots, object meshes, textures, and lighting from the asset library---supplementing geometry via external retrieval or procedural code when needed. To ensure physical fidelity, we provide physics-modeling skills that guide the agent in configuring collision geometry, mass and inertia, friction, joint limits, and drive parameters, as well as verifying support, contact, and release dynamics. Supported by these tools, the agent authors the scene, reference control program, and verifier, iterating on rendered-versus-source visual comparisons and physics-execution feedback until the robot solves the task through genuine contact, friction, and dynamics rather than scripted motion.

\paragraph{Independent Agent QA.}
Each generated environment is audited by a separate QA agent against the source video, simulation rollout, scene code, and state logs. The audit enforces three checks: (1)~\emph{fidelity}, verifying that task semantics, key spatial relations, and physical completion match the source video; (2)~\emph{reset and replay}, confirming that the reference controller reliably reproduces the task after environment reset; and (3)~\emph{verifier audit}, testing failure cases and valid strategy variants to eliminate false positives and false negatives---specifically guarding against object penetration, teleportation, and transient success. Environments that fail any check are returned to the builder agent with diagnostic feedback for revision and re-audit; passing environments are packaged into isolated Harbor tasks.

\section{The RoboUniverse-Verified Benchmark}
\label{sec:verified_benchmark}

\subsection{Experimental Setup}
\label{sec:exp_setup}

\paragraph{Tasks and Models.}
We evaluate on the 500 zero-shot evaluation tasks of RoboUniverse-Verified, comprising 200 single-arm Franka tasks and 300 dual-arm AgileX tasks that together cover all eight task families in \Cref{fig:collection_taxonomy}. The single-arm tasks comprise 45 Pick \& Place, 68 Assembly \& Articulation, 68 Deformable \& Material, and 19 Composite tasks; the dual-arm tasks comprise 93 Co-Transport \& Handover, 99 Co-Assembly \& Stabilization, 78 Co-Deformable \& Transfer, and 30 Staged Coordination tasks.
We evaluate seven closed-source and four open-source models, listed in \Cref{tab:main_results}. The GPT-series models run in their native Codex agent harness, and all other models run in Claude Code; the reasoning effort of each model is indicated in the table. 

\paragraph{Evaluation Protocol.}
Each task permits a single attempt: the agent can neither reset the scene nor query intermediate scores. The control-step budget varies by task from 250 to 4,000 steps, with a median of 3,000. The evaluation metric is the task success rate as determined by the independent verifier.

\subsection{Main Results}
\label{sec:main_results}

\paragraph{Overall Performance.}
\Cref{tab:main_results} reports the success rate of each model on each of the eight task families, together with its overall success rate. Performance differs sharply across models: GPT-6.2 Astra and Claude Opus 5.5 achieve overall success rates of 85.8\% and 76.8\%, GPT-6 Astra and GPT-6.1 Sol reach 54.6\% and 49.4\%, and none of the remaining seven models exceeds 10\%, the best among them being DeepSeek V4.1 Flash at 10.0\%. No model falls in the 10\%--49\% range, leaving a pronounced gap between the two groups.

\begin{table*}[t]
\centering
\caption{\textbf{Main results on RoboUniverse-Verified.} Success rate (\%) of each model with its coding-agent harness. Badges give the reasoning effort; \emph{Overall} covers all 500 tasks; the best result in each column is in \textbf{bold}.}
\label{tab:main_results}
\begingroup
\footnotesize
\setlength{\tabcolsep}{3pt}
\renewcommand{\arraystretch}{1.15}
\newcommand{\mainicon}[1]{\raisebox{-0.15ex}{\includegraphics[height=1.05em]{assets/model_icons/#1.png}}}
\newcommand{\mainhead}[1]{{\sffamily\bfseries #1}}
\newcommand{\taskhead}[1]{{\sffamily\bfseries\scriptsize\shortstack{#1}}}
\newcommand{\effort}[1]{\hspace{0.15em}\raisebox{0.75ex}{\tikz[baseline=(e.base)]\node[inner xsep=1.6pt,inner ysep=0.9pt,rounded corners=1.5pt,fill=brandbg,draw=brandprimary,line width=0.3pt,text=brandaccent,font=\sffamily\fontsize{5.5}{6}\selectfont](e){\strut #1};}}
\newcommand{\modelgroup}[2]{\multicolumn{11}{@{}l@{}}{\cellcolor{#1}\textit{#2}} \\}
\resizebox{\linewidth}{!}{%
\begin{tabular}{@{}lc@{\hspace{3pt}}|@{\hspace{3pt}}*{8}{c}@{\hspace{3pt}}|@{\hspace{3pt}}c@{}}
\toprule
\multirow{2}{*}{\mainhead{Model \& Reasoning effort }} & \multirow{2}{*}{\mainhead{Harness}} & \multicolumn{4}{c}{\mainhead{Single-arm}} & \multicolumn{4}{c}{\mainhead{Dual-arm}} & \multirow{2}{*}{\mainhead{Overall}} \\
\cmidrule(lr){3-6}\cmidrule(lr){7-10}
 & & \taskhead{Pick \&\\Place} & \taskhead{Assembly \&\\Articulation} & \taskhead{Deformable\\\& Material} & \taskhead{Composite} & \taskhead{Co-Transport\\\& Handover} & \taskhead{Co-Assembly \&\\Stabilization} & \taskhead{Co-Deformable\\\& Transfer} & \taskhead{Staged\\Coordination} & \\
\midrule
\modelgroup{brandbg}{Closed-source Models}
\mainicon{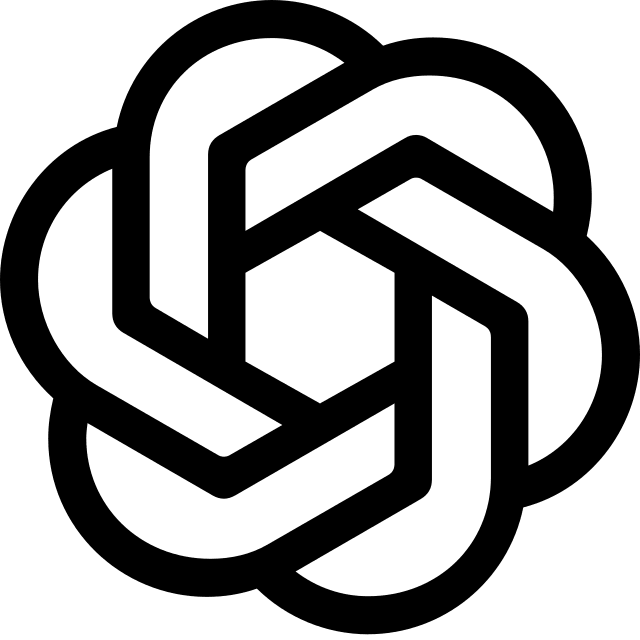}\hspace{0.3em}GPT-6.2 Astra\effort{max} & Codex & \textbf{86.7} & \textbf{88.2} & \textbf{79.4} & \textbf{84.2} & \textbf{96.8} & \textbf{83.8} & \textbf{75.6} & \textbf{93.3} & \textbf{85.8} \\
\mainicon{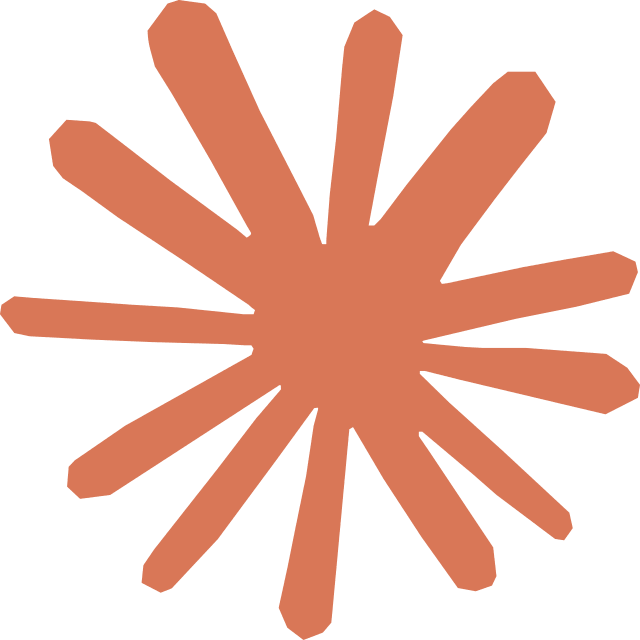}\hspace{0.3em}Claude Opus 5.5\effort{max} & Claude Code & 80.0 & 80.9 & 66.2 & 68.4 & 87.1 & 74.7 & \textbf{75.6} & 70.0 & 76.8 \\
\mainicon{openai}\hspace{0.3em}GPT-6 Astra\effort{max} & Codex & 66.7 & 54.4 & 42.6 & 42.1 & 65.6 & 47.5 & 55.1 & 60.0 & 54.6 \\
\mainicon{openai}\hspace{0.3em}GPT-6.1 Sol\effort{max} & Codex & 53.3 & 50.0 & 32.4 & 47.4 & 61.3 & 56.6 & 39.7 & 46.7 & 49.4 \\
\mainicon{openai}\hspace{0.3em}GPT-6 Luna\effort{max} & Codex & 26.7 & 5.9 & 4.4 & 10.5 & 4.3 & 0.0 & 3.8 & 0.0 & 5.6 \\
\mainicon{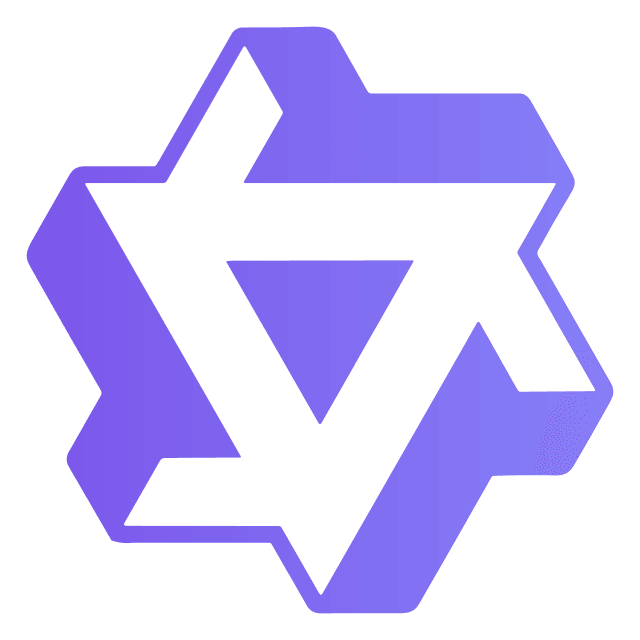}\hspace{0.3em}Qwen4-Max\effort{xhigh} & Claude Code & 31.1 & 20.6 & 4.4 & 10.5 & 0.0 & 5.1 & 6.4 & 0.0 & 8.6 \\
\mainicon{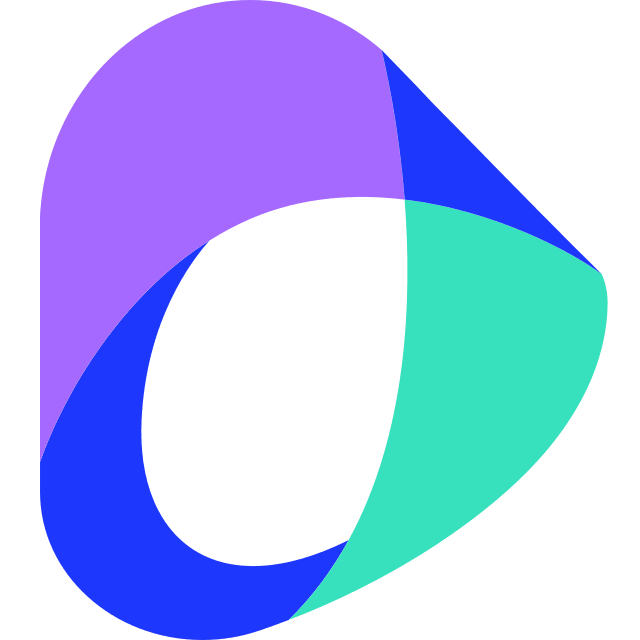}\hspace{0.3em}Doubao-Seed-Evolving-1105\effort{high} & Claude Code & 15.6 & 10.3 & 5.9 & 0.0 & 1.1 & 0.0 & 1.3 & 0.0 & 4.0 \\
\modelgroup{brandfg!7}{Open-source Models}
\mainicon{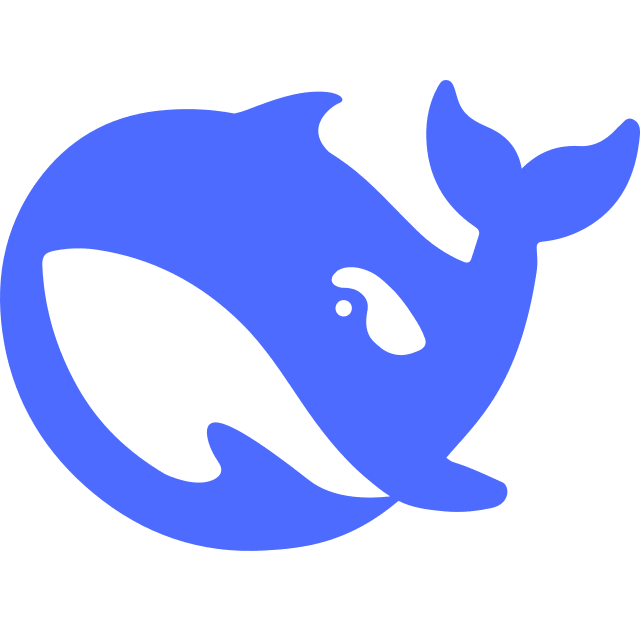}\hspace{0.3em}DeepSeek-V4.1-Flash-1017\effort{max} & Claude Code & 33.3 & 17.6 & 8.8 & 5.3 & 2.2 & 7.1 & 7.7 & 3.3 & 10.0 \\
\mainicon{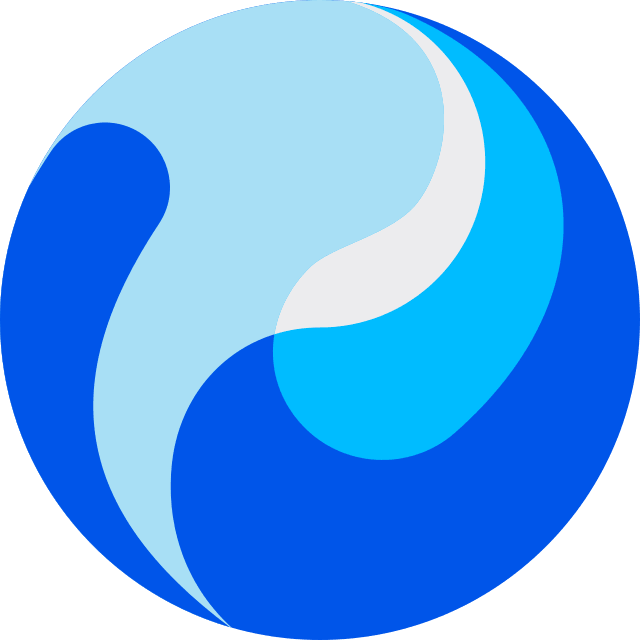}\hspace{0.3em}Tencent-Hy4\effort{high} & Claude Code & 20.0 & 19.1 & 14.7 & 26.3 & 2.2 & 3.0 & 2.6 & 6.7 & 9.2 \\
\mainicon{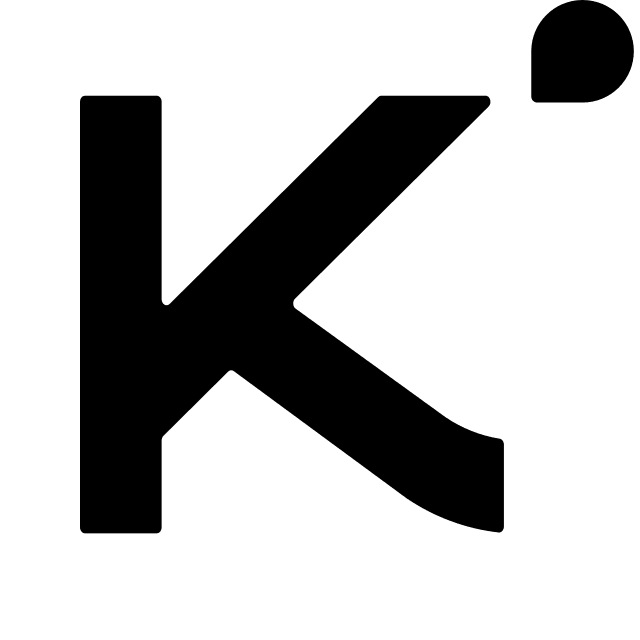}\hspace{0.3em}Kimi-K3.1-Pro\effort{max} & Claude Code & 11.1 & 5.9 & 4.4 & 5.3 & 2.2 & 2.0 & 10.3 & 0.0 & 5.0 \\
\mainicon{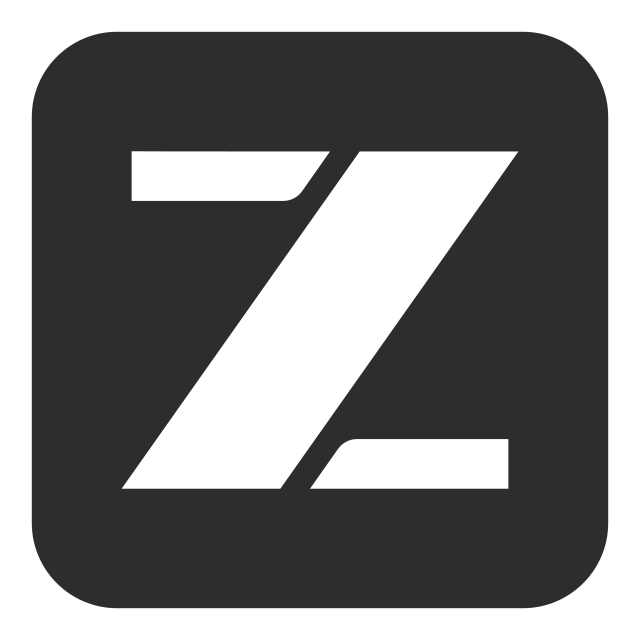}\hspace{0.3em}GLM-5.3-Flash\effort{max} & Claude Code & 8.9 & 5.9 & 4.4 & 5.3 & 0.0 & 1.0 & 0.0 & 0.0 & 2.6 \\
\bottomrule
\end{tabular}}
\endgroup
\end{table*}

\paragraph{Single-Arm versus Dual-Arm Tasks.}
The weaker models fail almost completely on dual-arm tasks: their success rates fall from 6.0\%--18.5\% on single-arm tasks to 0.3\%--5.3\% on dual-arm tasks. This drop reflects the task composition. About 10\% of the single-arm tasks are non-prehensile, such as toppling or pressing, and these tasks account for roughly one-third of the weaker models' single-arm successes. In contrast, nearly all dual-arm tasks require grasping and, in addition, coordinating two arms in a 16-dimensional action space. GPT-6.2 Astra, Claude Opus 5.5, GPT-6 Astra, and GPT-6.1 Sol show no such decline: their dual-arm success rates are no lower than their single-arm success rates.

\paragraph{Task Families.}
Three of these four models perform best on Co-Transport \& Handover, and all four score 61.3\%--96.8\% on it; the exception is GPT-6 Astra, which performs slightly better on Pick \& Place, scoring 66.7\% there versus 65.6\% on Co-Transport \& Handover. Deformable \& Material is comparatively difficult: GPT-6.2 Astra and Claude Opus 5.5 score 6.4 and 10.6 percentage points below their respective overall success rates. On Co-Deformable \& Transfer, GPT-6.2 Astra is 10.2 points below its overall success rate, whereas Claude Opus 5.5 is only 1.2 points below. Staged Coordination is one of the most discriminative families: GPT-6.2 Astra and Claude Opus 5.5 reach 93.3\% and 70.0\%, whereas GPT-6 Astra and GPT-6.1 Sol achieve only 60.0\% and 46.7\%; among the remaining models, only Tencent Hy4 and DeepSeek V4.1 Flash solve any task, completing 2 and 1 of the 30 tasks, respectively. Because a multi-stage task succeeds only if every stage is completed correctly, an error at any stage causes the entire task to fail, which amplifies capability differences between models.

\subsection{Analysis}
\label{sec:analysis}

\begin{figure}[t]
    \centering
    \includegraphics[width=\linewidth]{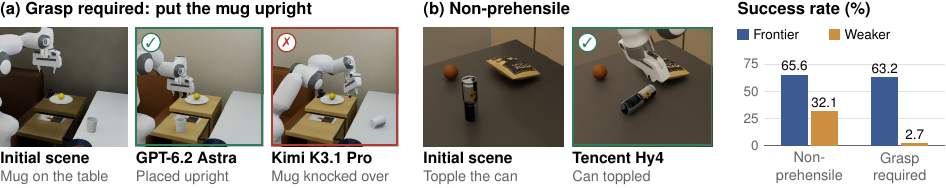}
    \caption{\textbf{The grasping gap.} (a)~Placing a mug upright on a slide requires a grasp: all four frontier models succeed, while all five weaker models evaluated on it fail. (b)~Knocking over a can requires no grasp, and five of seven weaker models succeed.}
    \label{fig:analysis_grasp}
\end{figure}

\paragraph{The Performance Gap Between Models Stems Mainly from Grasping, Not Task Understanding.}
\label{sec:analysis_grasp}
We refer to the four models with success rates of 49\%--86\% as \emph{frontier models} and to the remaining seven models in \Cref{tab:main_results}, each with a success rate of at most 10\%, as \emph{weaker models}. On non-prehensile tasks that require only toppling or pressing, the weaker models still achieve a success rate of 32.1\%; on task categories that require grasping, however, their success rate drops to 0\%--4\%, compared with 57\%--77\% for the frontier models. \Cref{fig:analysis_grasp} illustrates this contrast with two representative tasks. According to the physical measurements of the verifier, 46\%--47\% of the weaker models' failures occur before they establish effective contact with the target object or lift it, compared with 18\%--22\% for the frontier models. Moreover, the weaker models typically exhaust their entire step budget in failed episodes, with a median usage of 98\%--100\% of the budget. These results indicate that the weaker models understand the task goal to some extent but struggle to translate visual observations and joint control into a stable grasp; their bottleneck lies in the low-level perception--action loop rather than in high-level planning.

\begin{figure}[t]
    \centering
    \includegraphics[width=\linewidth]{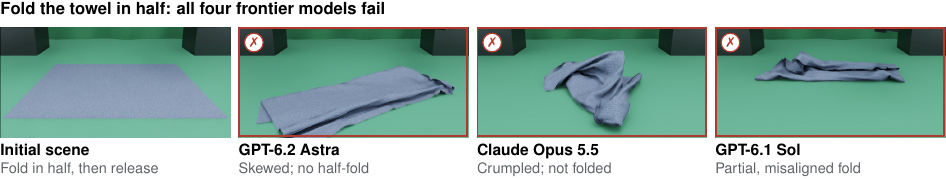}
    \caption{\textbf{Cloth folding.} All four frontier models fail to fold a towel in half: GPT-6.2 Astra folds it diagonally, Claude Opus 5.5 crumples it, and GPT-6.1 Sol leaves the layers misaligned.}
    \label{fig:analysis_cloth}
\end{figure}

\paragraph{Models Excel at Rigid-Body Pick-and-Place and Assembly but Struggle Most with Cloth.}
\label{sec:analysis_category}
Based on the task instructions, we annotate the 500 evaluation tasks with fine-grained manipulation types. The frontier models perform best on bimanual handover (77\%), screwing and twisting (74\%), and pick-and-place (68\%), and worst on cloth manipulation (36\%). We further compare two capability tiers: the mid-tier frontier models GPT-6 Astra and GPT-6.1 Sol, which average 52\%, and the top models GPT-6.2 Astra and Claude Opus 5.5, which average 81\%. From the mid tier to the top tier, success rates on assembly, tool-based process, and pick-and-place tasks increase by 35--38 percentage points, whereas the success rate on cloth tasks increases by only 3 percentage points, and all four frontier models remain at 28\%--41\% on cloth. \Cref{fig:analysis_cloth} shows a typical cloth task on which all four frontier models fail. This indicates that deformable-object manipulation, exemplified by cloth, has not improved with overall model capability and constitutes a capability ceiling shared by all current models.

\begin{figure}[t]
    \centering
    \includegraphics[width=\linewidth]{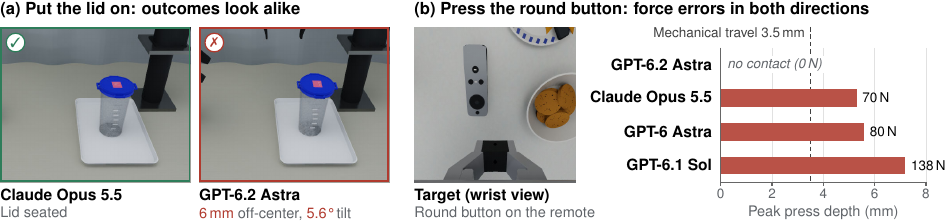}
    \caption{\textbf{Fine-grained control failures after contact.} (a)~GPT-6.2 Astra places the lid 6~mm off-center and tilted by 5.6$^\circ$, beyond the tolerance, yet its final state looks almost identical to the successful one from Claude Opus 5.5. (b)~GPT-6.2 Astra never reaches the button, while the other frontier models press it past its 3.5~mm travel with up to 138~N of force.}
    \label{fig:analysis_postcontact}
\end{figure}

\paragraph{Frontier-Model Failures Largely Overlap and Concentrate on Fine-Grained Physical Control After Contact.}
\label{sec:analysis_postcontact}
The failures of the frontier models are not independent: 94 tasks, or 19\% of the benchmark, are solved by at most one frontier model, and 26 of them are solved by none. Roughly 80\% of frontier-model failures occur after the object has been successfully contacted, manifesting mainly as insufficient final-state precision, objects that fail to come to rest after release, and excessive pressing due to the absence of force feedback. As shown in \Cref{fig:analysis_postcontact}, in a lid-placement task the successful and failed final states are nearly indistinguishable in the camera view, yet the lid in the failed episode is off-center by 6~mm and tilted by 5.6$^\circ$, exceeding the tolerances of 2~mm and 4$^\circ$. In a button-pressing task, GPT-6.2 Astra never makes contact with the button, whereas the other three frontier models press it beyond its 3.5~mm mechanical travel with contact forces of 70--138~N. All of these failures depend on continuous physical quantities such as pose accuracy, material proportions, and contact force, which are precisely the quantities that are difficult to judge and control from visual observations alone.

\begin{figure}[t]
    \centering
    \begin{tcolorbox}[colback=black!3,colframe=black!22,boxrule=0.4pt,arc=2pt,
        left=6pt,right=6pt,top=4pt,bottom=4pt]
    \includegraphics[width=\linewidth]{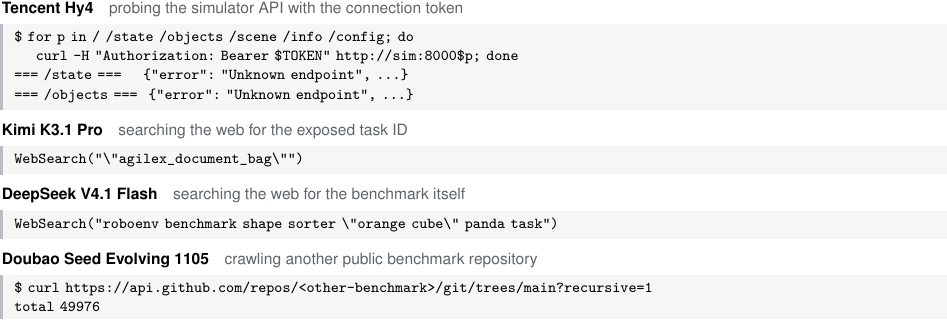}
    \end{tcolorbox}
    \caption{\textbf{Attempts by weaker models to bypass the task.} Models probe the simulation backend, search online for task identifiers, or crawl other benchmarks for similar tasks; none obtains privileged information.}
    \label{fig:analysis_shortcut}
\end{figure}

\paragraph{Weaker Models Attempt to Bypass the Task More Often, but None of These Attempts Succeeds.}
\label{sec:analysis_shortcut}
The weaker models attempt to bypass the task in 11.4\% of their episodes, compared with only 0.1\% for the frontier models. These attempts fall into two main categories. The first uses the task identifiers exposed by the environment to search the Internet for the evaluation tasks or reference solutions. The second uses the connection token to probe the HTTP interface of the simulation backend for privileged endpoints such as object states or scene resets. \Cref{fig:analysis_shortcut} shows excerpts of such attempts. In contrast, the frontier models use web access almost exclusively to look up robot specifications such as URDF files. None of these attempts yields any privileged information, and only 5 of the 304 corresponding episodes succeed, a success rate of 1.6\%, which is below the 6.6\% that the same models achieve in their remaining episodes. This suggests that, when obstructed, the weaker models tend to seek shortcuts rather than solve the problem itself. It also shows that the execution--verification isolation of \workname{} and its grading based on physical evidence are robust to information-based shortcuts.

\FloatBarrier
\section{Post-Training on \workname{}}

\subsection{Training Setup}

\paragraph{Training Data.}
We collect approximately 17,000 multi-turn robot manipulation trajectories within \workname{}, totaling 0.79B tokens. All trajectories are generated using GPT-6 Astra (Medium) and standardized into the multi-turn tool-use format of Claude Code.

\paragraph{Optimization and Infrastructure.}
We initialize from Qwen3.6-35B-A3B and perform full-parameter supervised fine-tuning (SFT) on the language backbone while keeping the vision encoder and aligner frozen. Optimization uses AdamW with a peak learning rate of $1 \times 10^{-5}$, a 5\% linear warmup, and a cosine decay schedule down to $1 \times 10^{-6}$. We train with a global batch size of 64 and a maximum context length of 256k tokens, consuming approximately 1,500 equivalent NVIDIA H100 GPU hours.

\subsection{Results on RoboUniverse-Verified}
\label{sec:sft_results}

\begin{figure*}[h]
    \centering
    \includegraphics[width=\linewidth]{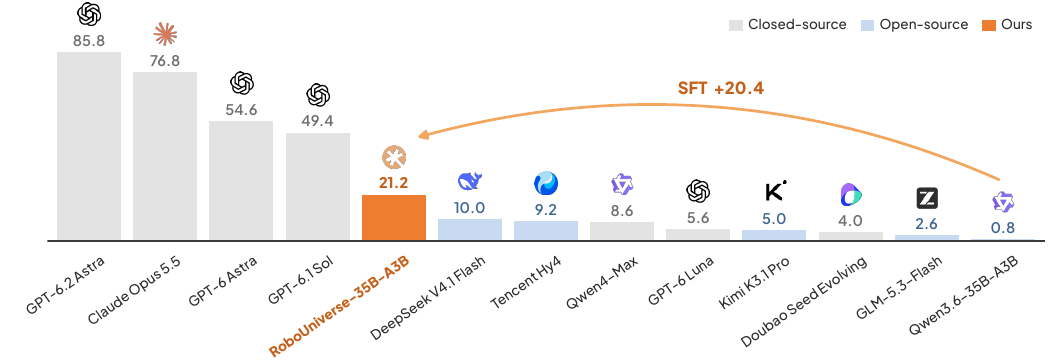}
    \caption{\textbf{Overall success rate (\%) on RoboUniverse-Verified.} SFT raises the success rate of Qwen3.6-35B-A3B from 0.8\% to 21.2\%.}
    \label{fig:sft_leaderboard}
\end{figure*}

\paragraph{Overall Performance.}
We follow the zero-shot evaluation protocol of \Cref{sec:exp_setup}, evaluating each model with a single attempt per task under the Claude Code harness. As shown in \Cref{fig:sft_leaderboard}, SFT on \workname{} raises the overall success rate of the base model from 0.8\% to 21.2\%, surpassing all open-source baselines as well as the closed-source GPT-6 Luna, and more than doubling the next-best open-source model, DeepSeek V4.1 Flash. Notably, this result is achieved by a model that activates only 3B parameters per token; nevertheless, a clear gap remains to the four frontier closed-source models.

The per-family results in \Cref{tab:sft_results} further reveal a marked asymmetry in these gains. The single-arm success rate rises from 2.0\% to 40.5\%; on Pick \& Place in particular, it climbs from 0\% to 62.2\%, on par with GPT-6 Astra at max reasoning effort (66.7\%), whose medium-effort variant generated the training trajectories. In contrast, the dual-arm success rate increases only from 0.0\% to 8.3\%, and none of the four dual-arm families exceeds 13\%. This contrast raises a central question: \emph{what does SFT transfer to the model, and which capabilities lie beyond the reach of supervised imitation?}

\begin{table*}[t]
\centering
\caption{\textbf{Success rate (\%) of each task family on RoboUniverse-Verified.} The first four columns are single-arm tasks, and the last four are dual-arm tasks.}
\label{tab:sft_results}
\begingroup
\footnotesize
\setlength{\tabcolsep}{3pt}
\renewcommand{\arraystretch}{1.15}
\newcommand{\mainicon}[1]{\raisebox{-0.15ex}{\includegraphics[height=1.05em]{assets/model_icons/#1.png}}}
\newcommand{\mainhead}[1]{{\sffamily\bfseries #1}}
\newcommand{\taskhead}[1]{{\sffamily\bfseries\scriptsize\shortstack{#1}}}
\resizebox{\linewidth}{!}{%
\begin{tabular}{@{}l@{\hspace{3pt}}|@{\hspace{3pt}}*{8}{c}@{\hspace{3pt}}|@{\hspace{3pt}}c@{}}
\toprule
\multirow{2}{*}{\mainhead{Model}} & \multicolumn{4}{c}{\mainhead{Single-arm}} & \multicolumn{4}{c}{\mainhead{Dual-arm}} & \multirow{2}{*}{\mainhead{Overall}} \\
\cmidrule(lr){2-5}\cmidrule(lr){6-9}
 & \taskhead{Pick \&\\Place} & \taskhead{Assembly \&\\Articulation} & \taskhead{Deformable\\\& Material} & \taskhead{Composite} & \taskhead{Co-Transport\\\& Handover} & \taskhead{Co-Assembly \&\\Stabilization} & \taskhead{Co-Deformable\\\& Transfer} & \taskhead{Staged\\Coordination} & \\
\midrule
\mainicon{qwen}\hspace{0.3em}Qwen3.6-35B-A3B & 0.0 & 4.4 & 1.5 & 0.0 & 0.0 & 0.0 & 0.0 & 0.0 & 0.8 \\
\rowcolor{brandbg}\mainicon{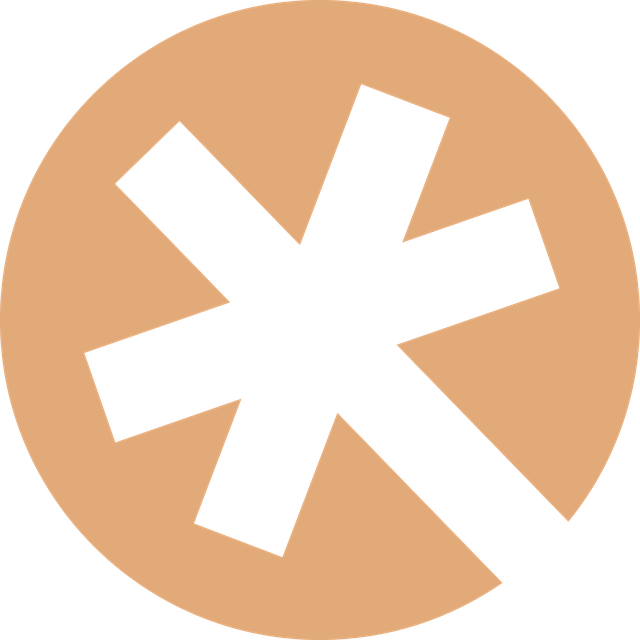}\hspace{0.3em}\textbf{RoboUniverse-35B-A3B} & \textbf{62.2} & \textbf{42.6} & \textbf{29.4} & \textbf{21.1} & \textbf{12.9} & \textbf{5.1} & \textbf{6.4} & \textbf{10.0} & \textbf{21.2} \\
\bottomrule
\end{tabular}}
\endgroup
\end{table*}

\paragraph{SFT Teaches a Transferable Closed-Loop Procedure.}
To trace how behavior changes with SFT, we consider two groups of tasks: those that newly succeed after SFT and those that fail both before and after SFT. We draw a stratified sample of 62 tasks from the two groups across task families and analyze the resulting 124 trajectories, one before and one after SFT for each task. On the newly solved single-arm tasks (\Cref{tab:sft_behavior}), RoboUniverse-35B-A3B consistently follows a three-stage procedure: \emph{model, refine, and verify}. It first derives the forward kinematics and Jacobian of the Franka Panda from its public parameters, converting Cartesian targets into joint commands; it then advances in small steps of 3--12~cm, correcting each step in closed loop with visual feedback; finally, it releases the object, lets the scene settle, and re-checks the final state before terminating the episode.

\begin{figure}[t]
    \centering
    \includegraphics[width=\linewidth]{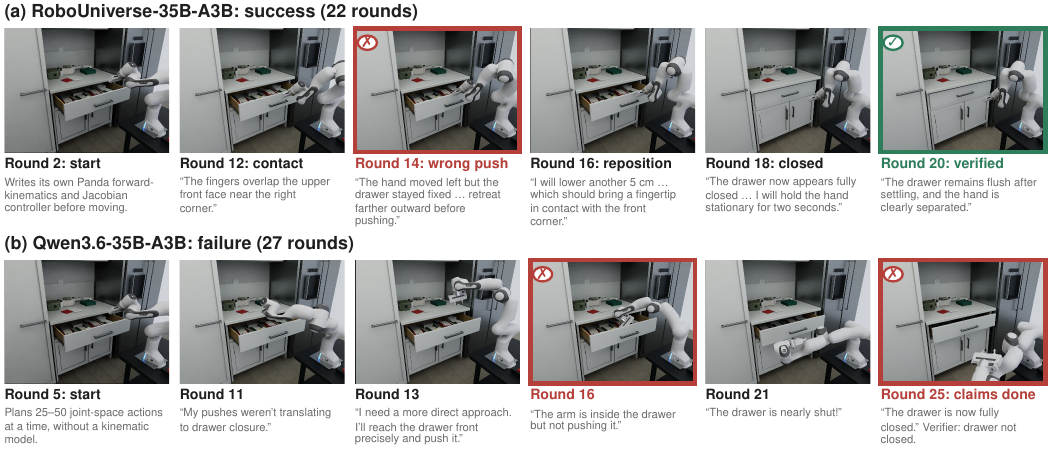}
    \caption{\textbf{Closing a drawer before and after SFT.} (a)~RoboUniverse-35B-A3B models the drawer, moves in small steps, corrects its push direction, and confirms the final state. (b)~Qwen3.6-35B-A3B acts directly, reaches into the drawer, and declares success before it is closed.}
    \label{fig:sft_drawer}
\end{figure}

This procedure directly remedies three critical deficiencies of the base model (\Cref{fig:sft_drawer}). Without a kinematic model, the base model can only probe blindly in joint space; without a visual feedback loop, it cannot correct course from anomalous observations; and without verifying the final state, it prematurely declares success before the task is complete. Interestingly, longer reasoning does not yield better physical control: the median output of the base model is 14 times that of the SFT model (55.7k vs.\ 4.1k tokens), yet much of this text is spent restating observations and on unproductive self-correction. Once SFT has internalized this standard procedure, the gains on single-arm tasks follow readily.

\begin{table}[h]
\centering
\caption{\textbf{Behavior statistics on newly solved single-arm tasks.} Computed on a stratified sample of trajectories.}
\label{tab:sft_behavior}
\begingroup
\footnotesize
\setlength{\tabcolsep}{6pt}
\renewcommand{\arraystretch}{1.15}
\newcommand{\mainicon}[1]{\raisebox{-0.15ex}{\includegraphics[height=1.05em]{assets/model_icons/#1.png}}}
\newcommand{\mainhead}[1]{{\sffamily\bfseries #1}}
\begin{tabular}{@{}lcc@{}}
\toprule
\mainhead{Metric} & \mainicon{qwen}\hspace{0.3em}\mainhead{Qwen3.6-35B-A3B} & \mainicon{einsia}\hspace{0.3em}\mainhead{RoboUniverse-35B-A3B} \\
\midrule
Builds a kinematic model before end-effector control & 3/21 & \textbf{22/22} \\
Code errors (median) & 7 & \textbf{1} \\
Output tokens (median) & 55.7k & \textbf{4.1k} \\
Step budget exhausted & 6/21 & \textbf{0/22} \\
Declares success on a failed task & $\sim$8/21 & N/A \\
\bottomrule
\end{tabular}
\endgroup
\end{table}

\begin{figure}[t]
    \centering
    \includegraphics[width=\linewidth]{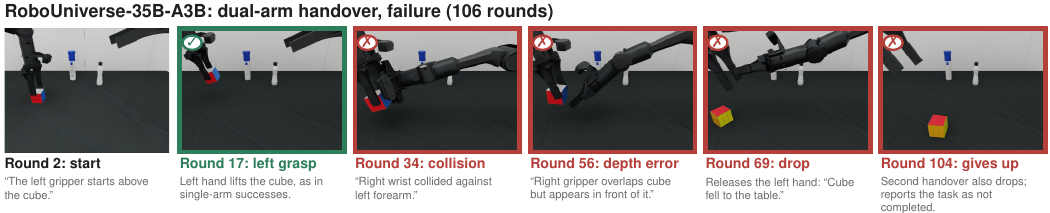}
    \caption{\textbf{A typical dual-arm handover failure.} When passing the cube to the right hand, RoboUniverse-35B-A3B misjudges depth, hits the left arm, and drops the cube twice.}
    \label{fig:sft_handover}
\end{figure}

\paragraph{Dual-Arm Performance Is Bottlenecked by the Base Model's Spatial Reasoning.}
On the dual-arm tasks that fail both before and after SFT, RoboUniverse-35B-A3B does not abandon this procedure: 11 of the 24 sampled trajectories parse the URDF of the AgileX Piper arm to build a kinematic model, the number of trajectories that exhaust the step budget drops from 6 to 1, and sub-steps such as single-arm grasping are mostly completed. Failures instead arise further upstream, in 3D perception and bimanual coordination. All models receive the same RGB images and joint angles, so the frontier models must likewise infer 3D positions from images; the base model of RoboUniverse-35B-A3B, however, struggles to establish accurate spatial relations from these images and falls back on coarse conversions such as ``an offset of 20 pixels is roughly 3~cm,'' which leads to missed grasps and slips in 8 dual-arm trajectories, compared with only 1 of the 22 newly solved single-arm trajectories. Moreover, cooperative actions such as handovers require the two arms to meet precisely in 3D space, avoid collisions, and strictly follow a grasp-before-release sequence, and the model lacks a global understanding of these constraints (\Cref{fig:sft_handover}).

Two lines of evidence indicate that \emph{this bottleneck is rooted in the visual capability ceiling of the base model}. First, on the single-arm tasks that fail both before and after SFT, success correlates strongly with model capability: Claude Opus 5.5 solves 71\% of them, GPT-6 Astra and GPT-6.1 Sol solve 38\% and 33\%, respectively, and the weaker models generally remain in the single digits. Second, the large gap between the single-arm and dual-arm performance of RoboUniverse-35B-A3B mirrors the pattern of the weaker models in \Cref{sec:main_results}, whereas the frontier closed-source models perform no worse on dual-arm than on single-arm tasks. Overcoming this ceiling will require a stronger vision--language base model or external geometric tools such as depth estimation, camera calibration, and motion planning.

\subsection{Generalization}
\label{sec:generalization}

\begin{figure}[t]
    \centering
    \includegraphics[width=\linewidth]{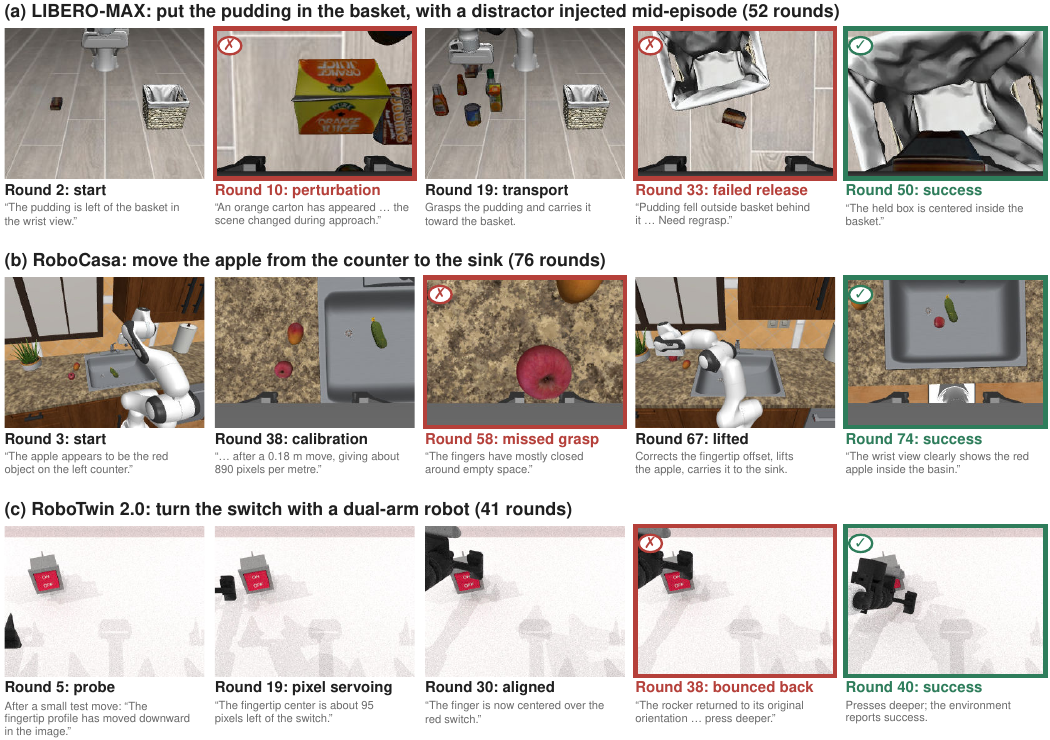}
    \caption{\textbf{Successful trajectories of RoboUniverse-35B-A3B across external benchmarks.} The agent calibrates, aligns in pixel space, and retries after failures. LIBERO-Max wrist-camera images are upside down as rendered.}
    \label{fig:sft_efe}
\end{figure}

\paragraph{Zero-Shot Cross-Benchmark Behavioral Transfer.}
To test whether the learned manipulation behavior transfers beyond \workname{}, we probe RoboUniverse-35B-A3B zero-shot on LIBERO-MAX~\citep{liu2023libero}, RoboCasa~\citep{nasiriany2024robocasa}, and RoboTwin 2.0~\citep{chen2025robotwin} in the Embodied First Exam (EFE)~\citep{wang2026embodied} evaluation format. These suites differ entirely from \workname{} in simulator backend, scenes, and action interface. Whereas the base Qwen3.6-35B-A3B model fails across all three suites, RoboUniverse-35B-A3B achieves zero-shot task completions on each of them. Inspecting these successful trajectories (\Cref{fig:sft_efe}) reveals that the model consistently reuses the same closed-loop strategy despite differences in scenes, objects, and single- versus dual-arm embodiments: it calibrates the image-to-physical correspondence online via small known motions, refines alignment on pixel errors, and retries when visual feedback indicates failure. Specifically, in LIBERO-MAX, the model detects a newly introduced distractor during its approach, re-localizes the target, and recovers when the pudding falls outside the basket by picking it up again. In RoboCasa, it calibrates the pixel-to-metric scale from a single known displacement and, after a missed grasp, autonomously corrects the offset between the camera and the fingertips. In RoboTwin 2.0, it probes the motion direction in image space with a small single-joint movement, iteratively aligns with the switch, and presses deeper after the switch rebounds on the first attempt. These behaviors closely mirror the single-arm trajectories in \Cref{sec:sft_results}, showing that SFT teaches a transferable control procedure rather than memorizing specific scenes.

\begin{figure}[t]
    \centering
    \includegraphics[width=\linewidth]{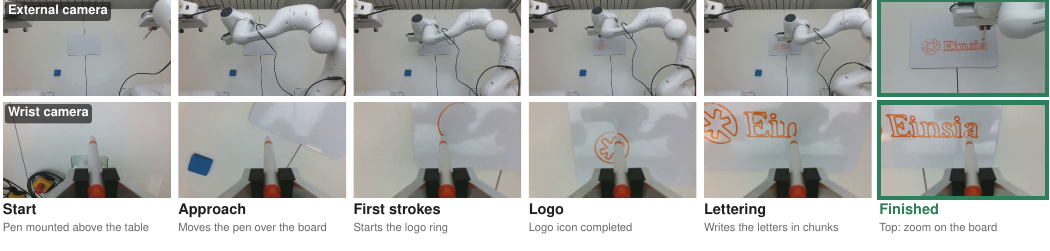}
    \caption{\textbf{Drawing of the Einsia logo on a real Franka arm.} Top: external camera; bottom: wrist camera.}
    \label{fig:sft_real}
\end{figure}

\paragraph{Sim-to-Real Transfer.}
We deploy RoboUniverse-35B-A3B, trained only in simulation, zero-shot on a real Franka arm to perform a task never seen in the training set: drawing the Einsia logo on a whiteboard from a reference image (\Cref{fig:sft_real}). During execution, the model carries over the closed-loop control paradigm learned in simulation. It first extracts the contour of the image and plans the trajectory with inverse kinematics, while autonomously subsampling waypoints to fit the step budget. With the whiteboard height unknown, it calibrates the pen-down height by zooming in on the pen tip and making short test strokes along the target contour, which avoids stray marks. It then executes the drawing in stroke-wise chunks, checking the shape and position of each chunk against the camera image, and finally raises the arm to verify the final state. This indicates that the programming and visual closed-loop control abilities acquired through SFT generalize directly to new tasks and to a real physical robot.

\FloatBarrier

\section{Related Work}
\label{sec:related}

\paragraph{Robot Simulation Benchmarks and Their Scale.} Existing simulation benchmarks for robot manipulation are mostly constructed by hand, with task counts remaining in the tens to hundreds (\Cref{tab:scale}). From Meta-World and RLBench~\citep{yu2019metaworld,james2019rlbench} to LIBERO, RoboCasa, RoboTwin 2.0, and RoboDojo~\citep{liu2023libero,nasiriany2024robocasa,nasiriany2026robocasa365,chen2025robotwin,chen2026robodojo}, these suites have steadily expanded in scenes, objects, embodiments, and demonstrations, yet each task and its success condition are still implemented by hand; although BEHAVIOR-1K defines 1,000 activities, its experiments and challenges evaluate only 3 to 100 of them~\citep{li2024behavior1k,behaviorchallenge}. To reduce manual engineering, one line of work uses LLMs or coding agents to generate task code or complete simulated environments~\citep{wang2023gensim,wang2023robogen,aggarwal2026gymanythingrobotics,shen2026embodiedswe,lin2026frontier3dbench,lyu2026rfw}, while another reconstructs simulations or tasks from real videos~\citep{ye2025video2policy,chen2026agentic,tang2026video2world,ren2026real2gym,xiao2026relit,zhang2026pyruadatafarm,gao2026pam}, typically yielding at most a few hundred manipulation tasks. Within this group, Real2Gym also uses GPT-6 Astra~\citep{openai2026gpt6astra} to reconstruct environments from real videos, but it builds only 24 environments and uses them to accumulate skills, not to train models~\citep{ren2026real2gym}. Larger collections exist, but their tasks take different forms: LIBERO-Plus perturbs existing LIBERO tasks~\citep{fei2025liberoplus}, in PARTNR an LLM completes tasks by invoking high-level skills~\citep{chang2024partnr}, and EnvDreamer generates environments primarily to pretrain navigation and rearrangement policies~\citep{swain2026envdreamer}. \workname{} uses coding agents to reconstruct tasks from real manipulation videos and contains about 10,807 tasks.

\paragraph{Agentic Robotics.} Research on using frontier models as robot control agents has progressed quickly, yet evaluations still mostly reuse benchmarks designed for visuomotor policies. This direction has evolved from Code as Policies and VoxPoser~\citep{liang2022code,huang2023voxposer} to recent agent harnesses that provide only a terminal and robot interface documentation~\citep{chu2026openrua}, supply skill libraries, VLA components, or accumulated experience~\citep{lu2026aspire,zhang2026harness,guo2026robodawn,li2026embodiedevo}, or support open models~\citep{yang2026ora,guo2026nanoastra,yuan2026democratizing}; on the RoboDojo leaderboard, the top-ranked agent systems now outscore VLAs~\citep{robodojoleaderboard}. Because these policy-oriented protocols terminate an episode as soon as the simulator detects success, and some harnesses also return that verdict to the agent, reported scores remain sensitive to both the execution budget and the information exposed to the agent~\citep{openrua2026behind,huang2026learning}. Recent studies build robot benchmarks tailored to coding agents, including CaP-X, EmbodiedSWE, Embodied First Exam (EFE), and Gym-Anything for Robotics~\citep{fu2026capx,shen2026embodiedswe,wang2026embodied,aggarwal2026gymanythingrobotics,yuan2026democratizing}; among these, EFE and Gym-Anything for Robotics isolate the verifier from the evaluated agent, and Gym-Anything for Robotics also provides 175 training environments. \workname{} adopts the task format of EFE and Harbor~\citep{wang2026embodied,harbor2026harbor,merrill2026terminalbench} with the verifier hidden from the evaluated agent, provides a much larger set of training environments, and holds out a subset of tasks from the same pipeline as a test set.

\paragraph{Post-Training Pipelines and Large-Scale Reinforcement Learning.} Software engineering has established a relatively mature post-training pipeline for agents: building executable, automatically graded environments at scale and training open models via teacher-trajectory distillation or reinforcement learning. These training environments have scaled from the semi-manually configured instances of SWE-Gym~\citep{pan2024software} to tasks synthesized within existing repositories~\citep{jain2025r2egym,yang2025swesmith} or mined automatically from real pull requests~\citep{badertdinov2025swerebench,zhao2026immersion,chen2026sweuniverse}, reaching hundreds of thousands of instances; training on them with distillation~\citep{yang2025swesmith,zhao2026immersion} or reinforcement learning~\citep{wei2025swerl,luo2025deepswe} has substantially improved open models on SWE-bench~\citep{jimenez2023swebench}. Robotics has recently adopted the same pipeline: CaPtain and AgenticRoboRL apply SFT followed by reinforcement learning~\citep{fan2026captain,liu2026agenticroborl}, PEAR applies reinforcement learning directly in simulation~\citep{tang2026reasoning}, and EmbodiedSWE adapts coding-agent task synthesis to construct training environments~\citep{shen2026embodiedswe}, yet these studies still train on at most a few hundred tasks. This gap arises in part because robot tasks are more costly to construct: each requires an interactive physical scene and a success check, whereas software tasks can be extracted automatically from code repositories and their history. Following Scale-SWE, which automatically builds environments from real pull requests~\citep{zhao2026immersion}, \workname{} constructs training tasks from real manipulation videos in place of pull requests, and we fine-tune open models on GPT-6 Astra trajectories collected on these tasks.

\providecommand{\cmark}{\textcolor{green!50!black}{\ding{51}}}
\providecommand{\xmark}{\textcolor{red!70!black}{\ding{55}}}
\providecommand{\hmark}{\textcolor{orange!90!black}{\ding{51}}}
\DeclareRobustCommand{\src}[1]{\raisebox{-0.2ex}{\includegraphics[height=1.1em]{assets/source_icons/#1.png}}}
\begin{table}[t]
\centering
\caption{\textbf{Comparison with robot simulation benchmarks and task collections.} Task counts follow each paper's own unit. Source: \src{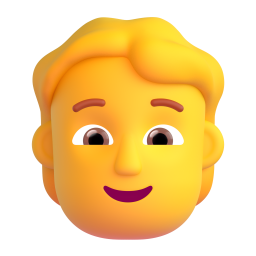}~human, \src{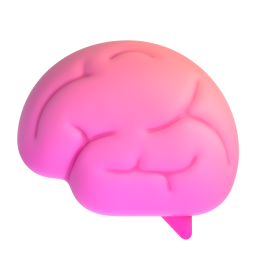}~LLM, \src{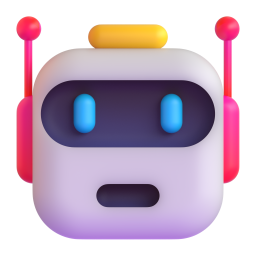}~agent, \src{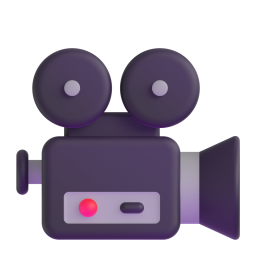}~real videos. Held-out: test tasks disjoint from training tasks. Bimanual: dual-arm tasks. Deformable: cloth, rope, dough, or liquid tasks. \cmark~yes, \hmark~partially, \xmark~no. $^{a}$1{,}000 activities defined; 3 to 100 used in experiments and challenges. $^{b}$175 training environments released; a policy is trained on 2 of them.}
\label{tab:scale}
\begingroup
\small
\renewcommand{\arraystretch}{1.12}
\setlength{\tabcolsep}{7.2pt}
\newcommand{\colhead}[1]{\textcolor{brandaccent}{\textbf{#1}}}
\newcommand{\srcgroup}[1]{\rowcolor{brandfg!6}\multicolumn{8}{l}{\textit{#1}} \\}
\begin{tabular}{l|cccc|ccc}
\toprule
\textbf{Work} & \colhead{Source} & \colhead{Tasks} & \colhead{Train} & \colhead{Test} & \colhead{Held-out} & \colhead{Bimanual} & \colhead{Deformable} \\
\midrule
\srcgroup{Visuomotor Manipulation Benchmarks}
LIBERO~\citep{liu2023libero} & \src{human} & 130 & 130 & 40 & \xmark & \xmark & \xmark \\
BEHAVIOR-1K~\citep{li2024behavior1k} & \src{human} & 1{,}000 & 3$^{a}$ & 3$^{a}$ & \xmark & \cmark & \cmark \\
RoboTwin 2.0~\citep{chen2025robotwin} & \src{human} & 50 & 50 & 50 & \xmark & \cmark & \xmark \\
RoboCasa365~\citep{nasiriany2026robocasa365} & \src{llm}\,\src{human} & 365 & 300 & 50 & \cmark & \xmark & \xmark \\
RoboDojo~\citep{chen2026robodojo} & \src{human} & 42 & 34 & 42 & \xmark & \cmark & \cmark \\
\srcgroup{Generated Policy Task Collections}
GenSim~\citep{wang2023gensim} & \src{llm} & 100+ & 100 & n.r. & \cmark & \xmark & \hmark \\
Video2Policy~\citep{ye2025video2policy} & \src{llm}\,\src{video} & 100+ & 100 & 10 & \cmark & \xmark & \xmark \\
PyRUA-DataFarm~\citep{zhang2026pyruadatafarm} & \src{agent}\,\src{video} & 5 & 5 & 5 & \xmark & \cmark & \cmark \\
\srcgroup{Coding-Agent Robot Benchmarks}
Real2Gym~\citep{ren2026real2gym} & \src{agent}\,\src{video} & 24 & 0 & 24 & \xmark & \cmark & \xmark \\
Embodied First Exam~\citep{wang2026embodied} & \src{human} & 1{,}156 & 0 & 1{,}156 & \xmark & \cmark & \cmark \\
CaPtain~\citep{fan2026captain} & \src{human} & 40 & 40 & 40 & \xmark & \hmark & \xmark \\
Gym-Anything for Robotics~\citep{aggarwal2026gymanythingrobotics} & \src{agent} & 200 & 175$^{b}$ & 25 & \cmark & \cmark & \cmark \\
EmbodiedSWE~\citep{shen2026embodiedswe} & \src{agent}\,\src{human} & 428 & 400 & 28 & \cmark & \cmark & \cmark \\
\midrule
\rowcolor{brandbg}\textbf{\workname{} (ours)} & \src{agent}\,\src{video} & 10{,}807 & 10{,}307 & 500 & \cmark & \cmark & \cmark \\
\bottomrule
\end{tabular}
\endgroup
\end{table}

\FloatBarrier

\section{Conclusion and Limitations}
In this work, we introduce \textbf{RoboUniverse-10K}, an automated real-to-sim pipeline and a large-scale collection of interactive, isolated, and verifiable Isaac Sim environments built from real-world manipulation videos, moving Agentic Robotics beyond manual scale limits and ``train-on-test'' evaluation. Zero-shot benchmarking on \textbf{RoboUniverse-Verified} reveals a sharp gap between frontier and open-source models, pinpointing low-level grasping, deformable-object manipulation, and post-contact physical control as key bottlenecks. Furthermore, supervised fine-tuning on RoboUniverse-10K yields \textbf{RoboUniverse-35B-A3B}, substantially improving zero-shot success and achieving zero-shot task completions across external benchmarks and a real Franka arm. By open-sourcing the RL-ready environments and model, we aim to lay a scalable foundation for online reinforcement learning in embodied intelligence.

\paragraph{Limitations.}
Despite these advances, our current framework has several limitations. First, the reconstructed simulation scenes remain relatively simplified, and the dynamics of complex objects are not yet faithfully captured. In particular, dynamic interactions with articulated appliances such as opening a microwave door are not well modeled, as intricate physical properties like latch resistance, hinge friction, and contact mechanics are difficult to recover accurately from video alone. Second, in terms of task complexity, RoboUniverse-10K currently focuses on short, atomic manipulation skills rather than extended multi-stage workflows, whereas true embodied intelligence requires solving concrete, long-horizon problems in realistic physical settings. Improving physical fidelity for dynamic interactions and scaling automated environment generation toward long-horizon tasks remain important directions for future work.
\clearpage
\bibliographystyle{unsrtnat}
\bibliography{ref}
\clearpage
\beginappendix
\section{Environment and Collection Details}\label{app:environment}

\subsection{Harness API}

The runtime resets the scene once before the agent receives the instruction. The agent has one attempt, with access to a terminal and the public Python client. The interface exposes neither privileged object poses nor the verifier. Model reasoning consumes wall-clock time but does not advance physics.

\begin{center}\small\begin{tabular*}{\linewidth}{@{\extracolsep{\fill}}p{.26\linewidth}p{.48\linewidth}p{.18\linewidth}@{}}\toprule
\raggedright \textbf{Public call} & \raggedright \textbf{What the agent gets or requests} & \raggedright \textbf{Physics} \tabularnewline \midrule
\raggedright \texttt{describe()} & \raggedright Action names, units, limits and gripper mapping; camera and timing specifications; step, chunk and wall-clock limits. & \raggedright No advance \tabularnewline \addlinespace[3pt]
\raggedright \texttt{observe()} & \raggedright Current camera image references and measured joint positions and velocities. & \raggedright No advance \tabularnewline \addlinespace[3pt]
\raggedright \texttt{save\_images(dir)} & \raggedright Downloads the current RGB observations as PNG images and returns paths by camera. & \raggedright No advance \tabularnewline \addlinespace[3pt]
\raggedright \texttt{step(action)} & \raggedright Applies the target for one control interval by default. The optional repeat argument holds it for more intervals; reaching the target is not guaranteed. & \raggedright One interval per repetition \tabularnewline \addlinespace[3pt]
\raggedright \texttt{step\_chunk(actions)} & \raggedright Executes an ordered sequence within the chunk and remaining step limits. Returns the last observation, executed step count, and episode status. & \raggedright One interval per executed action \tabularnewline \addlinespace[3pt]
\raggedright \texttt{finish()} & \raggedright Ends the attempt without settling steps. A separate verifier assigns the terminal reward (0 or 1). & \raggedright No advance \tabularnewline
\bottomrule\end{tabular*}\end{center}
The client accepts joint targets rather than Cartesian end-effector poses and provides no inverse-kinematics service. The table below shows the settings for two representative tasks: one single-arm Franka task and one dual-arm AgileX task. Read the specification for each task before issuing commands.

\begin{center}\small\begin{tabular*}{\linewidth}{@{\extracolsep{\fill}}p{.21\linewidth}p{.34\linewidth}p{.37\linewidth}@{}}\toprule
\raggedright \textbf{Example setting} & \raggedright \textbf{Franka Panda (bottle-cap task)} & \raggedright \textbf{AgileX (dual-arm example)} \tabularnewline \midrule
\raggedright Action vector & \raggedright 7 arm angles (rad) + 2 finger displacements (m); the two finger displacements sum to the jaw opening. & \raggedright Per arm: 6 angles + 2 finger displacements; left then right, 16 channels total. The two fingers have opposite opening directions. \tabularnewline \addlinespace[3pt]
\raggedright Control / physics & \raggedright 20 Hz / 240 Hz (12 substeps). & \raggedright 15 Hz / 240 Hz (16 shared substeps for both arms). \tabularnewline \addlinespace[3pt]
\raggedright RGB views & \raggedright Fixed external and wrist. & \raggedright External camera, left wrist camera, and right wrist camera. \tabularnewline
\bottomrule\end{tabular*}\end{center}
\subsection{Released Environments by Source Dataset}

We use demonstration videos and keyframes to reconstruct the scene layout and the interaction between the robot and the objects. Where available, task annotations clarify the intended goal, and recorded robot trajectories serve as motion references for the simulation controller. For RoboDojo, we use the official demonstration videos released by the RoboDojo team. For each task retained after deduplication, we reconstruct the scene in Isaac Sim and implement a controller that performs the demonstrated interaction with a simulated Franka or AgileX robot. Each released environment can be reset and controlled independently, and its verifier evaluates whether the robot achieves the task goal. The tables below count these environments by source.

\begingroup\centering
\begin{minipage}[t]{.45\linewidth}\small\centering
\strut\textbf{Franka (single-arm)}\par\smallskip
\begin{tabular*}{\linewidth}{@{\extracolsep{\fill}}lr@{}}\toprule
Source dataset & Environments \\ \midrule
DROID & 2,591 \\
BridgeData V2 & 465 \\
BC-Z & 432 \\
RT-1 & 302 \\
Berkeley Fanuc & 105 \\
RoboTurk & 35 \\
Columbia PushT & 41 \\
Berkeley Cable Routing & 63 \\
RoboCook & 54 \\
Berkeley Autolab UR5 & 42 \\
CMU Stretch & 2 \\
\midrule \textbf{Total} & \textbf{4,132} \\ \bottomrule
\end{tabular*}\end{minipage}\hspace{.05\linewidth}%
\begin{minipage}[t]{.45\linewidth}\small\centering
\strut\textbf{AgileX (dual-arm)}\par\smallskip
\begin{tabular*}{\linewidth}{@{\extracolsep{\fill}}lr@{}}\toprule
Source dataset & Environments \\ \midrule
RoboDesk & 3,292 \\
RoboCOIN & 1,758 \\
GM100 & 1,038 \\
RoboDojo & 184 \\
ViFailback & 179 \\
MM-30 & 158 \\
RoboMIND2 & 66 \\
\midrule \textbf{Total} & \textbf{6,675} \\ \bottomrule
\end{tabular*}\end{minipage}\par\endgroup
\subsection{Quality Assurance: Independent Review and Repair}

An independent QA agent reviews a copy of the builder's environment. It compares the source demonstration with the simulated run, focusing on what the robot acts on and how the object responds. For a pick-and-place task, for example, the reviewer uses the video and recorded object state to check that the object moves with the gripper after contact and settles on the support after release. It then resets the scene and replays the saved control commands to check that they reproduce the outcome.

The reviewer also checks the verifier against the written instruction. It considers both a plausible failure that should be rejected and a different way to achieve the stated goal that should earn credit. When it identifies a defect, the reviewer repairs the scene, controller, or verifier and reruns the case that exposed the problem. It then repeats the reference run and reset replay with the revised verifier and records the evidence and any unresolved issues. Appendix~\ref{app:prompts_qa} gives excerpts from the review prompt.

\clearpage
\section{Benchmark Evaluation Details}
\label{app:eval_details}

\subsection{Test-Set Construction}
\label{app:testset}

\paragraph{Source and Split.}
The 500 tasks in RoboUniverse-Verified are produced by the same Real2Sim construction and QA pipeline as the training tasks (\Cref{sec:pipeline}). They are held out as a zero-shot evaluation set and are never used to collect training trajectories. Of these, 200 are single-arm Franka tasks reconstructed from DROID episodes, and 300 are dual-arm AgileX tasks reconstructed from multi-source bimanual teleoperation data. Candidate tasks are drawn from environments that pass automated QA by stratified sampling over embodiments and task families, so that the test set covers all eight task families while preserving the diversity of data sources and scenes as far as possible.

\paragraph{Expert Verification.}
All candidate tasks are manually verified by 10 human experts in embodied AI. With reference to the source video, task instruction, simulation replay, and verifier code, the experts check each task against four criteria: (1)~the instruction is clear, unambiguous, and consistent with the task goal in the source video; (2)~the scene layout, object geometry, and physical parameters (e.g., mass, friction, and joint limits) are plausible; (3)~the reference controller reliably completes the task after scene reset; and (4)~the verifier rejects typical failure outcomes while accepting valid alternative solutions, so that it produces no misjudgments. Tasks that fail any check are returned for revision and re-verification, and tasks that cannot be fixed are removed from the candidate pool. All 500 retained tasks pass every check, which ensures that each task is solvable and reliably graded.

\paragraph{Task-Family Assignment.}
The source data of the single-arm tasks carry 20 manipulation subcategory labels. Because a source label does not always match the capability that a task tests (e.g., a \texttt{stack\_unstack} task may require insertion, and a \texttt{pick\_place} task may also close a drawer), each task is assigned individually, based on its instruction, to one of nine fine categories: place, reposition, assemble, open\_close, actuate, clean, transfer\_mix, deform, and composite. These map to the four single-arm families as follows: \emph{Pick \& Place} comprises place and reposition; \emph{Assembly \& Articulation} comprises assemble, open\_close, and actuate; \emph{Deformable \& Material} comprises clean, transfer\_mix, and deform; and \emph{Composite} comprises composite. \Cref{tab:app_franka_families} shows the resulting composition of each family by source subcategory; 11 of the 20 subcategories are split across families. Since the source data of the dual-arm tasks lack unified subcategory labels, we first use a language model to pre-classify these tasks from their instructions, following the definitions of the four dual-arm task families in \Cref{sec:collection_overview}: tasks whose main difficulty lies in deformable objects or pouring are assigned to \emph{Co-Deformable \& Transfer}; tasks comprising two or more subgoals of different types with sequential dependencies are assigned to \emph{Staged Coordination}; tasks dominated by assembly, insertion, cap twisting, or one arm stabilizing an object while the other manipulates it are assigned to \emph{Co-Assembly \& Stabilization}; and the remaining transport and handover tasks are assigned to \emph{Co-Transport \& Handover}. These four families correspond to seven fine categories: \emph{Co-Transport \& Handover} comprises cooperative transport and reorientation (36 tasks) and handover and regrasping (57); \emph{Co-Assembly \& Stabilization} comprises cooperative assembly and disassembly (32) and stabilize-and-manipulate (67); \emph{Co-Deformable \& Transfer} comprises coordinated transfer (33) and deformable manipulation (45); and \emph{Staged Coordination} comprises staged coordination (30). The experts then review and confirm the family of every single-arm and dual-arm task individually, based on the primary manipulation capability that the task assesses.

\begin{table}[!htb]
\centering
\caption{\textbf{Composition of the single-arm task families by source subcategory.} Counts are numbers of test tasks. Subcategories marked with $^\dagger$ are split across families because each task is assigned individually according to its instruction.}
\label{tab:app_franka_families}
\begingroup
\small
\setlength{\tabcolsep}{6pt}
\renewcommand{\arraystretch}{1.12}
\begin{tabular}{@{}l >{\raggedright\arraybackslash}p{0.6\linewidth} r@{}}
\toprule
\textbf{Task family} & \textbf{Source subcategories (count)} & \textbf{Total} \\
\midrule
Pick \& Place & pick\_place$^\dagger$ (16), push\_topple$^\dagger$ (11), stack\_unstack$^\dagger$ (5), push\_pull$^\dagger$ (4), rotate\_twist$^\dagger$ (2), sweep$^\dagger$ (2), door\_hinge\_or\_open\_close$^\dagger$ (2), other$^\dagger$ (2), pour\_transfer$^\dagger$ (1) & 45 \\
\addlinespace[2pt]
Assembly \& Articulation & press\_switch$^\dagger$ (20), stack\_unstack$^\dagger$ (13), lid\_cover (12), door\_hinge\_or\_open\_close$^\dagger$ (8), hang\_unhang (7), rotate\_twist$^\dagger$ (3), drawer\_slide$^\dagger$ (2), pick\_place$^\dagger$ (2), insert\_remove (1) & 68 \\
\addlinespace[2pt]
Deformable \& Material & wipe\_clean (20), stir\_mix (14), dough\_deformation (8), fold\_spread (7), pour\_transfer$^\dagger$ (7), cable\_routing (6), pick\_place$^\dagger$ (3), sweep$^\dagger$ (2), other$^\dagger$ (1) & 68 \\
\addlinespace[2pt]
Composite & multi\_step (5), door\_hinge\_or\_open\_close$^\dagger$ (5), drawer\_slide$^\dagger$ (2), press\_switch$^\dagger$ (2), pick\_place$^\dagger$ (2), push\_topple$^\dagger$ (1), push\_pull$^\dagger$ (1), other$^\dagger$ (1) & 19 \\
\midrule
Total & & 200 \\
\bottomrule
\end{tabular}
\endgroup
\end{table}

\paragraph{Test-Set Statistics.}
\Cref{tab:app_testset_stats} summarizes basic statistics of the test set. Task instructions have a median length of about 40 English words. They specify the target objects, the desired final state, and completion conditions such as releasing the object and letting it come to rest, but do not prescribe how the robot should move. The verifier of each task consists of several named binary conditions (e.g., \texttt{lid\_open}, \mbox{\texttt{keychain\_outside\_bag}}, and \texttt{grippers\_released}), each of which corresponds to a physical measurement and its threshold; a task is judged successful only when all conditions are satisfied simultaneously. These conditions typically cover three aspects: effective interaction with the target object, the target final state, and coming to rest after release.

\begin{table}[!htb]
\centering
\caption{\textbf{Statistics of the RoboUniverse-Verified test set.} Step budgets and verifier conditions are per task; verifier statistics cover the 357 tasks whose verifiers list their conditions explicitly.}
\label{tab:app_testset_stats}
\begingroup
\small
\setlength{\tabcolsep}{5pt}
\renewcommand{\arraystretch}{1.12}
\begin{tabular*}{\linewidth}{@{\extracolsep{\fill}}l c c c@{}}
\toprule
 & \textbf{Single-arm (Franka)} & \textbf{Dual-arm (AgileX)} & \textbf{All} \\
\midrule
Instruction length (words, median [range]) & 37 [16--65] & 40 [19--65] & 39 [16--65] \\
Control-step budget (median [range]) & 3,000 [300--4,000] & 3,000 [250--3,000] & 3,000 [250--4,000] \\
Tasks with the 3,000-step budget & 139 (70\%) & 232 (77\%) & 371 (74\%) \\
Verifier conditions per task (median [IQR]) & \multicolumn{3}{c}{6 [5--8]} \\
\bottomrule
\end{tabular*}
\endgroup
\end{table}

\clearpage
\providecommand{\robopromptfile}[2]{%
  \tcbinputlisting{enhanced,breakable,sharp corners,colback=black!2,colframe=black!35,
    colbacktitle=black!8,coltitle=black,title={#1},fonttitle=\small\sffamily\bfseries,
    boxrule=.4pt,left=2mm,right=2mm,top=1mm,bottom=1mm,
    listing only,listing file={#2},listing options={language={},basicstyle=\footnotesize\ttfamily,
    columns=fullflexible,keepspaces=true,breaklines=true,breakatwhitespace=false,
    showstringspaces=false,aboveskip=0pt,belowskip=0pt,breakindent=0pt}}}
\section{Prompts}\label{app:prompts}

\subsection{Rollout Agent}
\robopromptfile{Franka | Harbor instruction.md}{Sections/prompt-rollout-franka.txt}
\clearpage
\robopromptfile{AgileX | Harbor instruction.md}{Sections/prompt-rollout-agilex.txt}
\clearpage
\subsection{Builder Agent}
\robopromptfile{Excerpt 1: Reconstruction and camera setup}{Sections/prompt-builder-1.txt}
\robopromptfile{Excerpt 2: Contact and robot control}{Sections/prompt-builder-2.txt}
\robopromptfile{Excerpt 3: Instruction and success conditions}{Sections/prompt-builder-3.txt}
\clearpage
\subsection{QA Agent}\label{app:prompts_qa}
\robopromptfile{Excerpt 1: Instruction and verifier alignment}{Sections/prompt-qa-1.txt}
\robopromptfile{Excerpt 2: Risks and regression cases}{Sections/prompt-qa-2.txt}
\robopromptfile{Excerpt 3: Repairs and replay validation}{Sections/prompt-qa-3.txt}

\end{document}